\documentclass[review]{elsarticle}

\usepackage{lineno,hyperref}
\usepackage[normalem]{ulem} 
\usepackage{times}
\usepackage{epsfig}
\usepackage{graphicx}
\usepackage{amsmath}
\usepackage{amssymb}

\usepackage{mathtools}
\usepackage{gensymb}
\usepackage{subcaption}
\usepackage{url}
\usepackage{multicol}
\usepackage{multirow}
\usepackage{pdfpages}

\usepackage{array}
\usepackage{multirow}
\usepackage{textcomp}
\usepackage{array}
\usepackage{colortbl}

\modulolinenumbers[5]

\newcommand{\ie}{{\it i.e.},~}
\newcommand{\etal}{{\it et al.}~}
\renewcommand{\ie}{{\it i.e.},~}
\newcommand{\eg}{{\it e.g.},~}

\definecolor{myblue}{rgb}{0,0,1}
\definecolor{myred}{rgb}{0.8, 0, 0}
\definecolor{mygreen}{rgb}{0, 0.6, 0}

\newcommand{\cellgreen}[1]{\cellcolor{green!45} {#1}}

\graphicspath{{./graphics/}}

\journal{Image and Vision Computing Journal}

\begin{document}

\begin{frontmatter}

\title{Post-mortem Iris Recognition with Deep-Learning-based Image Segmentation}

\author[addressNASK]{Mateusz Trokielewicz\corref{correspondence}}

\cortext[correspondence]{Corresponding author}
\ead{mateusz.trokielewicz@nask.pl}

\author[addressND]{Adam Czajka}

\author[addressWUM]{Piotr Maciejewicz}

\address[addressNASK]{Biometrics and Machine Intelligence Laboratory, Research and Academic Computer Network\\Kolska 12, 01-045 Warsaw, Poland}
\address[addressND]{Department of Computer Science, University of Notre Dame\\46556 IN, USA}
\address[addressWUM]{Department of Ophthalmology, Medical University of Warsaw\\Lindleya 4, 02-005 Warsaw, Poland}

\begin{abstract}
This paper proposes the first known to us iris recognition methodology designed specifically for post-mortem samples. We propose to use deep learning-based iris segmentation models to extract highly irregular iris texture areas in post-mortem iris images. We show how to use segmentation masks predicted by neural networks in conventional, Gabor-based iris recognition method, which employs circular approximations of the pupillary and limbic iris boundaries. As a whole, this method allows for a significant improvement in post-mortem iris recognition accuracy over the methods designed only for ante-mortem irises, including the academic OSIRIS and commercial IriCore implementations. The proposed method reaches the EER less than 1\% for samples collected up to 10 hours after death, when compared to 16.89\% and 5.37\% of EER observed for OSIRIS and IriCore, respectively. For samples collected up to 369 hours post-mortem, the proposed method achieves the EER 21.45\%, while 33.59\% and 25.38\% are observed for OSIRIS and IriCore, respectively. Additionally, the method is tested on a database of iris images collected from ophthalmology clinic patients, for which it also offers an advantage over the two other algorithms. This work is the first step towards post-mortem-specific iris recognition, which increases the chances of identification of deceased subjects in forensic investigations. The new database of post-mortem iris images acquired from 42 subjects, as well as the deep learning-based segmentation models are made available along with the paper, to ensure all the results presented in this manuscript are reproducible.    
\end{abstract}


\end{frontmatter}


\section{Introduction}
\label{sec:Intro}
\subsection{Post-mortem iris recognition}
Post-mortem iris recognition has recently gained attention from the biometrics community, starting with a Master's thesis from Boston University \cite{BostonPostMortem}, and studies on \emph{ex vivo} pig eyes in 2015 \cite{PostMortemPigs}. Since then, some research groups have been hard at work evaluating how this biometrics works with irises of deceased subjects \cite{TrokielewiczPostMortemICB2016, TrokielewiczPostMortemBTAS2016, BolmePostMortemBTAS2016, Sauerwein_JFO_2017, TrokielewiczTIFS2018}, and proposing PAD (Presentation Attack Detection) techniques aimed at detecting cadaver iris presentations to the sensor \cite{TrokielewiczBTASPAD2018, ColdPAD_MITReview2018}. Post-mortem iris recognition is also discussed as a potentially useful method for forensic proceedings in addition to typical forensic biometric characteristics such as fingerprints, if the method could be applied reliably \cite{TrokielewiczTIFS2018}. Despite this attention, concluded with many claims that post-mortem iris biometrics, however challenging, may still be possible in favorable conditions, no specific, cadaver-aware methods for improving the recognition reliability have been proposed, apart from the work of Trokielewicz \etal \cite{TrokielewiczIWBF2018}, who introduced a data-driven semantic segmentation model for iris localization within images collected from cadaver eyes, yet without evaluating how the new segmentation algorithm improves the recognition accuracy.

\subsection{{Forensic motivation}}
{There are several scenarios in which post-mortem iris recognition can be useful due to its speed (when compared to usually slower DNA analysis), and which have been already explicitly appreciated by forensics community.

One of them is more accurate matching of ante-mortem samples (if available) with post-mortem data obtained at crime scenes and mass fatality incidents. One of the United States Department of Justice's operational requirements is defined as: ``Enhancement of unidentified decedent system(s) with weighting capability for antemortem and postmortem comparisons with the goal of providing a ranked list of “best matches” to effectively and efficiently identify potential candidates or hits.''\footnote{\url{https://www.nij.gov/topics/forensics/documents/2018-2-forensic-twg-table.pdf}}. The work presented in this paper directly addresses this requirement.

The second, operational practice-driven application, is to rapidly register the body at the scene, to track it later and dispatch it correctly to family or mortuary. In case of mass fatality accidents, DNA is too slow and forensic practitioners already decided to use iris recognition, which has been demonstrated for the first time in 2015 in Sansola's thesis defended at the Boston Medical School\footnote{\url{https://open.bu.edu/handle/2144/13975}}.}

\subsection{Contributions of this work}
In this work, we build upon the findings presented in \cite{TrokielewiczIWBF2018}, by training several additional models with larger amounts of post-mortem data and different ground truth mask creation rules, together with constructing an iris image normalization method based on the circular Hough transform (CHT). This end-to-end segmentation algorithm is then coupled with the well known open source iris recognition method OSIRIS \cite{OSIRIS}, and evaluated against the baseline OSIRIS performance and against the commercial IriCore method, which presented the best performance of post-mortem iris matching in past studies \cite{TrokielewiczTIFS2018}. {\bf The contributions that this study makes towards the state-of-the-art in post-mortem iris recognition are thus the following:}
\begin{itemize}
	\item introduction of two new, deep convolutional neural network-based (DCNN-based) post-mortem iris image segmentation methods, trained with different ground truths, namely coarse iris approximation, and fine-grained, post-mortem specific masking of the regions affected by tissue decay, 
	\item an end-to-end iris localization and image segmentation model that can be used as a drop-in replacement for any iris recognition methodology that uses normalized images with iris boundary approximated by circles,
 	\item a new dataset of cadaver iris images collected from 42 subjects over a time period of up to 369 hours post-mortem\footnote{{Release agreement for this dataset is available at \url{http://zbum.ia.pw.edu.pl/EN/node/46}}},
	\item experiments showing a considerable improvement in the matching performance of a method employing the proposed segmentation and Gabor wavelet-based encoding,
	\item source codes and neural network models' weights, which allow for full reproducibility of the results, as well as incentive for further research.
\end{itemize}         

This article is laid out as follows. After the introductory part, Sec. \ref{sec:Related} discusses the historical assumptions and the current course of post-mortem iris recognition research. Biometric databases used for training and estimation, as well as evaluation of the proposed approach are introduced in Sec. \ref{sec:Database}, whereas in Sec. \ref{sec:Methods} we describe the methodology of our solution and ways for its evaluation, which is performed and reported in Sec. \ref{sec:experiments}. Finally, Sec. \ref{sec:Conclusions} contains relevant conclusions.

\section{Related work}
\label{sec:Related}  
Even before any meaningful experimental studies regarding post-mortem iris recognition had come to fruition, both the research community and the industry were skeptical about the feasibility of post-mortem iris recognition \cite{DaugmanPostMortem, IrisGuardPostMortem, IriTechPostMortem}.

The first systematic study in this field was Sansola's MSc thesis at Boston University \cite{BostonPostMortem}, in which the IriShield M2120U iris recognition camera and IriCore matching software were employed to photograph irises of 43 deceased subjects at different post-mortem time intervals. The method yielded 19-30\% of false non-matches and no false matches, depending on the time since death involved. Later on, Saripalle \etal \cite{PostMortemPigs} studied ex-vivo eyes of domestic pigs and their biometric capability during the degradation of tissues after they are taken out of the cadaver. They found that the irises lose their biometric capabilities 6 to 8 hours after death, proving the fast rate of ex-vivo eye degradation.

In our previous works, we have refuted most of the popular claims, showing that the iris can still successfully serve as a biometric identifier for 27 hours after death \cite{TrokielewiczPostMortemICB2016}, with the pupil typically remaining mid-sized without any excessive dilation or constriction, with well visible iris structure. More than 90\% of the irises were correctly recognized when photographed a few hours after death, including 100\% accuracy obtained for the IriCore method (which comes from the same vendor as the camera used in this study). As time after death increases, the equal error rate increases to 13.3\% when images captured approximately 27 hours after death are compared against those obtained 5h after demise. Later, we showed that correct matches can still be expected even after 17 days \cite{TrokielewiczPostMortemBTAS2016} and offered the first known to us database of 1330 NIR and VIS post-mortem iris images acquired from 17 cadavers \cite{WarsawColdIris1}. 

A study tracking biometric capabilities of both the typically used forensic modalities, namely fingerprints and faces, as well as iris is presented by Bolme \etal \cite{BolmePostMortemBTAS2016}. The authors placed twelve deceased subjects in outdoor conditions to assess how the environment and time affect the biometric performance. Irises were found to be degrading very fast and become unusable as shortly as few days after the body placement. When the bodies were kept outside for 14 days, the correct verification rate reported by the authors decreased to 0.6\%, but the real-life chance of recognizing an iris is estimated to be less than 0.1\%. Fingerprints and faces, on the other hand, remained readable for much longer periods of time.

 Sauerwein \etal \cite{Sauerwein_JFO_2017} showed that irises stay readable for up to 34 days after death, when cadavers were kept in outdoor conditions during winter. This readability parameter, however, was assessed by human experts acquiring the samples and no automatic iris recognition algorithms were used.

The most recent paper in this study by Trokielewicz \etal \cite{TrokielewiczTIFS2018} included the most comprehensive experiments in this field to date, and introduces the new dataset of cadaver iris images \cite{WarsawColdIris2}, which in combination with the first part of the Warsaw corpus aggregates 1,200 near-infrared and 1,787 visible light samples collected from 37 deceased subjects kept in mortuary conditions. In the foretold paper, we employed four iris recognition methods to analyze the genuine and impostor comparison scores and evaluate the dynamics of iris decay over a period of up to 814 hours. The findings show that iris recognition may be close-to-perfect 5 to 7 hours after death and occasionally is still viable even 21 days after death. We also demonstrated that false-match probability is significantly higher when live iris images are compared with post-mortem samples than when only live samples are used in comparisons.

\section{Databases of iris images}
\label{sec:Database}
For the purpose of this work, we use four different databases of iris images, including three subject-disjoint sets of post-mortem iris images, namely the:
\begin{itemize}
	\item {\bf Warsaw-BioBase-Postmortem-Iris-v1.1}, comprising 574 near-infrared (NIR) and 1023 visible light (VIS)  images collected from 17 cadavers over a period of up to 34 days \cite{WarsawColdIris1, TrokielewiczPostMortemBTAS2016}, 
	\item {\bf Warsaw-BioBase-Postmortem-Iris-v2}, a subject-disjoint extension of the v1.1 of the database, adding 626 NIR and 764 VIS data from 20 more cadavers \cite{WarsawColdIris2, TrokielewiczTIFS2018},
	\item {\bf Warsaw-BioBase-Postmortem-Iris-v3}, a new set of images collected for the purpose of this study, adding data from 42 more subjects with a total of 1094 NIR images and 785 VIS images, collected over a time horizon of up to 369 hours since demise.
\end{itemize}

The first two datasets are aggregated together and used as training data, whereas the third set is employed for evaluation purposes. This way, the training and testing data are subject-disjoint, \ie data from subjects `participating' in the training of the segmentation models and estimation of the normalization algorithm's parameters is not repeated in the evaluation experiments.  

In addition to the post-mortem data, we use another database of challenging iris images for supplemental testing of the proposed approach:
\begin{itemize}
	\item {\bf Warsaw-BioBase-Disease-Iris-v1}, comprising iris images collected mostly from elderly patients of an ophthalmology clinic, including subjects with various ophthalmic conditions; we use a subset of 551 NIR images from 76 randomly chosen eyes \cite{WarsawDiseaseIris1, TrokielewiczCYBCONF2015}.
\end{itemize}  

\begin{figure}[h!]
\centering
\includegraphics[width=0.49\linewidth]{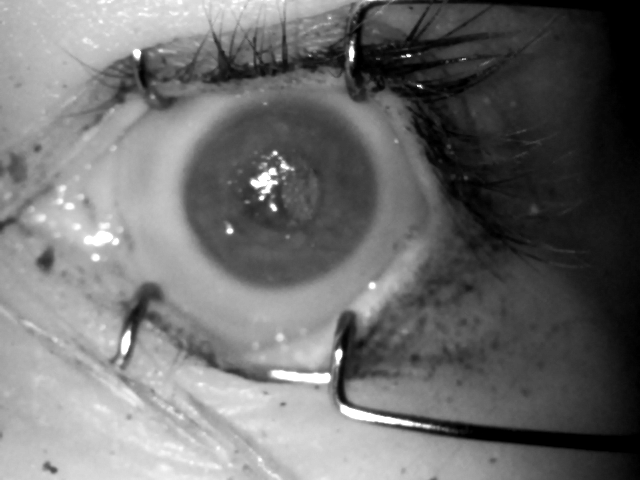}\hskip1mm
\includegraphics[width=0.49\linewidth]{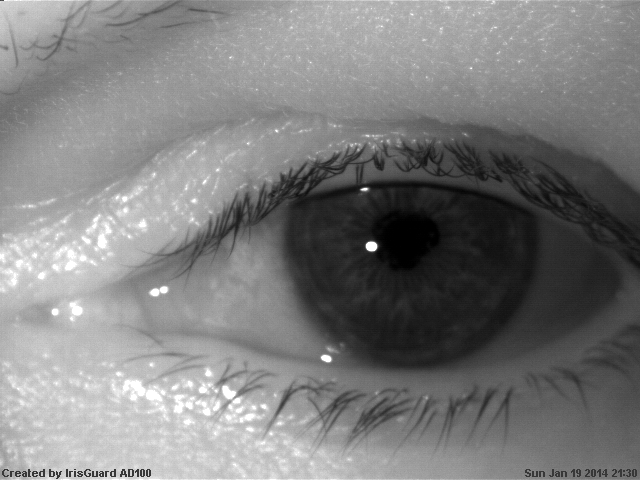}\hskip1mm
\caption{Example images from the test databases used in this work. Warsaw-BioBase-Postmortem-Iris-v3 ({\bf left}) and Warsaw-BioBase-Disease-Iris-v1 ({\bf right}.)}
\label{fig:testsets:examples}
\end{figure}

Following the findings reported in \cite{TrokielewiczTIFS2018} that strongly favor the NIR images for post-mortem iris recognition purposes, we only employ this type of images in all experiments conducted in this work. Example images from each of the evaluation databases are shown in Fig. \ref{fig:testsets:examples}. 

\section{Proposed methodology}
\label{sec:Methods}
\subsection{Baseline iris recognition method}
\label{sec:OSIRISintro}
OSIRIS, or \emph{Open Source for IRIS}, is an academic solution developed as a part of the BioSecure EU project \cite{OSIRIS} and implements the original iris recognition concept of John Daugman, including iris image segmentation and normalization to dimensionless polar coordinate system, followed by the calculation of a binary iris code using phase quantization of the Gabor wavelet filtering outcomes. Fractional Hamming distance calculated using the XOR operation between the two codes is used as a comparison score, with values close to zero expected for two same-eye images, whereas scores oscillating around 0.5 should be produced when two different-eye images are compared. Due to the additional compensation of the eyeball rotation, different-eye score distributions are typically skewed toward 0.4 -- 0.45 range.

{The OSIRIS, although the last version was developed in 2013, is still considered by many researchers in the biometric community as the most accurate {\bf open source} iris matcher. It also closely follows a dominant pipeline in iris recognition. We are not aware of any other open-source iris methods that would outperform OSIRIS. The Masek's implementation\footnote{\url{https://www.peterkovesi.com/studentprojects/libor/index.html}} and the USIT toolkit\footnote{\url{http://www.wavelab.at/sources/Rathgeb16a/}} performed worse than OSIRIS on post-mortem iris samples in our research, and thus we decided to not include them.}

The original score calculation method can be complemented by score normalization penalizing comparisons with small number of compared bits, as compared to the \emph{typical} (here: within a given dataset) number of compared bits, as proposed by Daugman \cite{Daugman2007NewMethods}:
$$HD_{norm} = 0.5 - (0.5 - HD_{raw}) \sqrt{\frac{n}{N}}$$
\noindent
This transforms the samples of scores obtained when comparing different eyes into samples drawn from the same binomial distribution, as opposed to drawing sample scores from different binomial distributions with $\sigma$ dependent on the number of bits $n$ that were available for comparison (commonly unmasked bits). $N$ is the `typical number of bits compared (unmasked) between two different irises,' which Daugman states to be 911 or 960, depending on the data. The $N$ parameter is estimated for a particular database of iris images. Since this work uses different datasets of iris images and different subsets of these datasets (\ie post-mortem samples with varying time horizon), the $N$ is calculated separately for each experiment (cf. description in Sec. \ref{sec:experiments}). The total number of bits in the OSIRIS code is 1536.

In the following Sections we will evaluate the proposed approach and discuss whether applying this normalization procedure is beneficial for the overall system's performance. 

\subsection{Data-driven image segmentation models}
The first stage of the proposed scheme consists of employing a data-driven segmentation model based on a deep convolutional neural network (DCNN) for the purpose of localizing the iris within an image. Our models are based on the solution described in one of our previous papers \cite{TrokielewiczIWBF2018}, which employs a re-trained off-the-shelf SegNet model \cite{SegNet2016}, and offers a large improvement over the OSIRIS segmentation method measured with the IoU (Intersection over Union) metric. We fine-tuned it with a database of cadaver iris images and their corresponding ground truth binary masks. {The neural networks were trained and tested with entirely subject-disjoint sets of iris images. Overfitting was minimized in a standard way by applying cross-validation with the training subset. The network generalizes well on images collected from subjects unknown during the training.} The masks used for training were annotated in a fine-grained manner, which has a purpose of teaching the network to only localize iris regions that are not affected by post-mortem changes, and represent the well-visible iris texture.

\begin{figure}[h!]
\centering
{\bf OSIRIS} segmentation, normalized images, and normalized masks\\\vskip1mm
\includegraphics[width=0.4\linewidth]{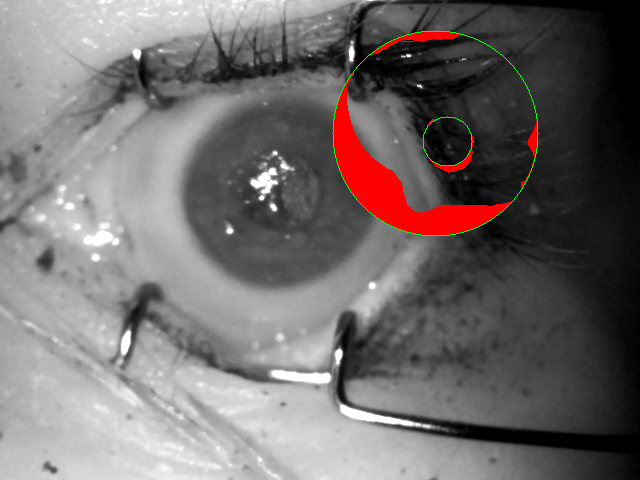}\hskip1mm
\includegraphics[width=0.4\linewidth]{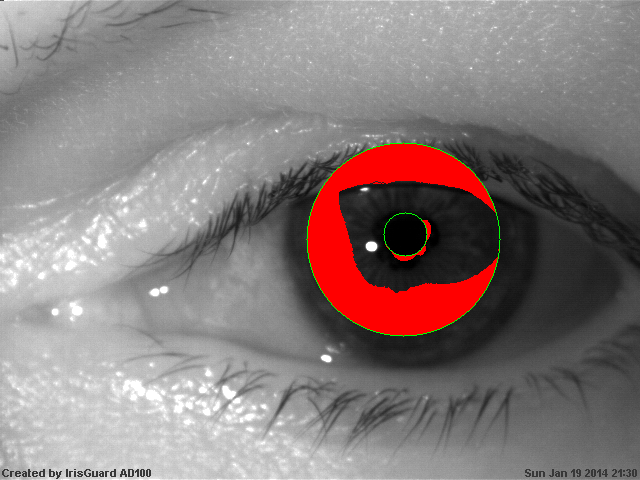}\\\vskip1mm
\includegraphics[width=0.4\linewidth]{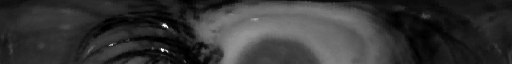}\hskip1mm
\includegraphics[width=0.4\linewidth]{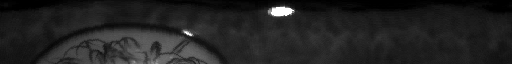}\\\vskip1mm
\includegraphics[width=0.4\linewidth]{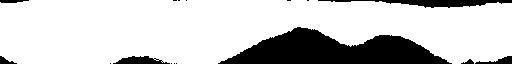}\hskip1mm
\includegraphics[width=0.4\linewidth]{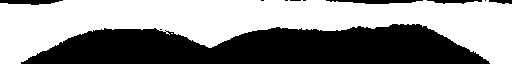}\\
{\it \textbf{coarse}} \textbf{CNN} binary predictions with fitted Hough circles, segmented images, normalized images, and normalized masks\\\vskip1mm
\includegraphics[width=0.4\linewidth]{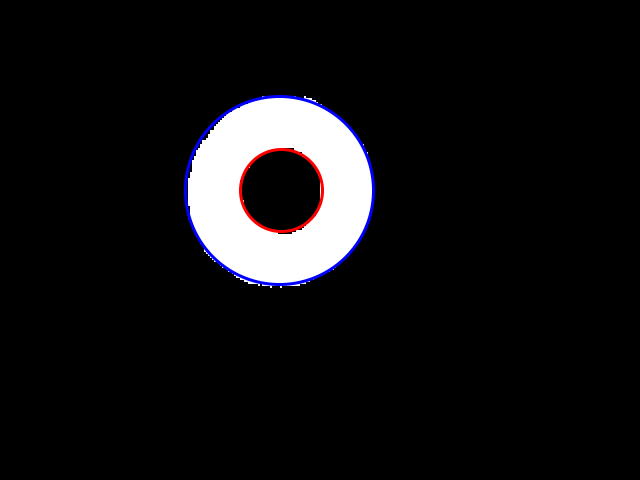}\hskip1mm
\includegraphics[width=0.4\linewidth]{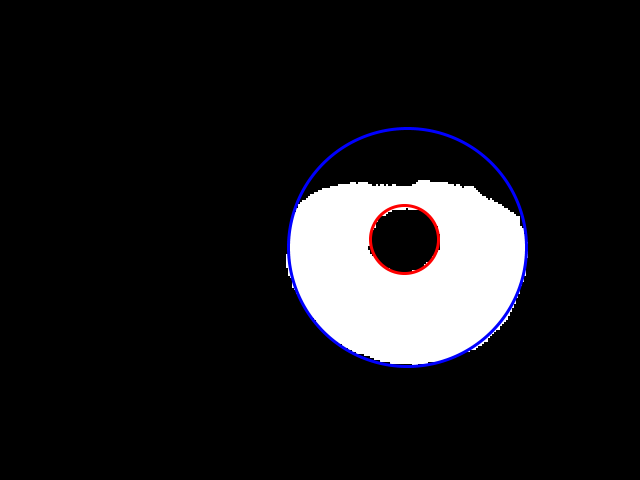}\\\vskip1mm
\includegraphics[width=0.4\linewidth]{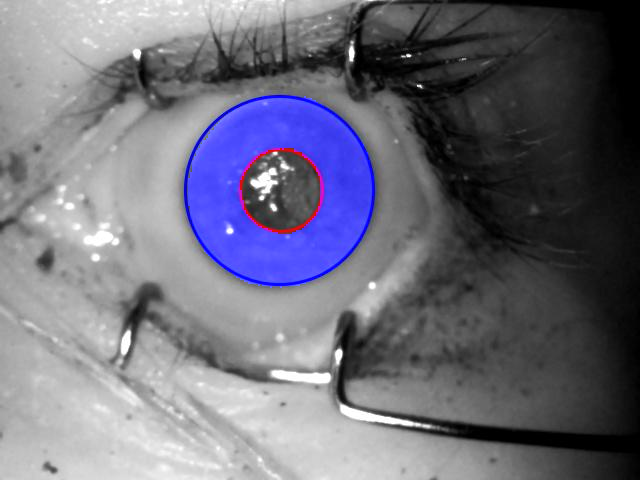}\hskip1mm
\includegraphics[width=0.4\linewidth]{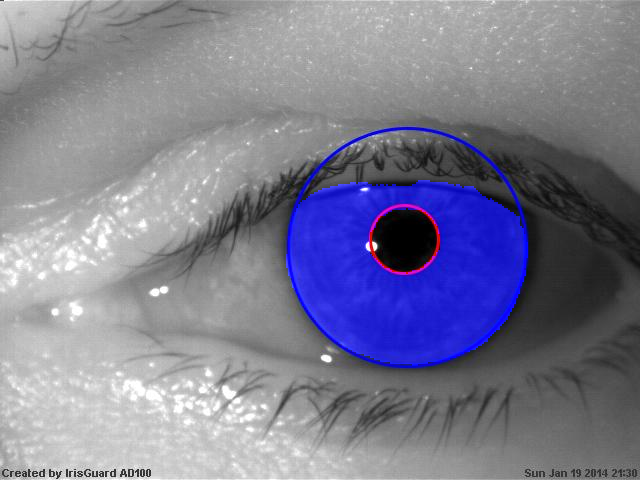}\\\vskip1mm
\includegraphics[width=0.4\linewidth]{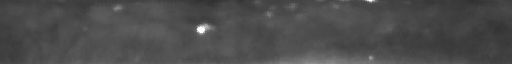}\hskip1mm
\includegraphics[width=0.4\linewidth]{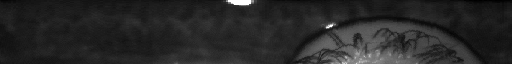}\\\vskip1mm
\includegraphics[width=0.4\linewidth]{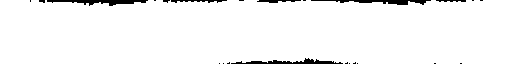}\hskip1mm
\includegraphics[width=0.4\linewidth]{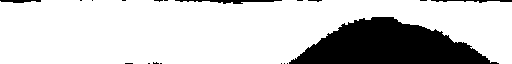}
\caption{Segmentation results for samples from Fig. \ref{fig:testsets:examples}, obtained from the unmodified OSIRIS method and the {\it \textbf{coarse}} segmentation model. Warsaw-BioBase-Postmortem-Iris-v3 ({\bf left}), Warsaw-BioBase-Disease-Iris-v1 ({\bf right}).}
\label{fig:models:examples}
\end{figure}

In addition to the model described above, in this work we have implemented and trained two additional models, ending up with three models in total, which are detailed below and their example predictions given in Fig. \ref{fig:models:examples}
\begin{enumerate}
	\item {\it \textbf{fine}} segmentation model: the one published in \cite{TrokielewiczIWBF2018}, trained with data from the Warsaw-Biobase-Postmortem-Iris-v1.1 database with \textbf{fine-grained ground truth masks} and yielding predictions of $120\times160$ pixels,
	\item {\it \textbf{fine v2highres}} segmentation model: similar to the \emph{fine} model, but trained with data from Warsaw-Biobase-Postmortem-Iris-v1.1 database together with NIR samples from the Warsaw-Biobase-Postmortem-Iris-v2 database for twice as many epochs (120 vs 60), also with \textbf{fine-grained ground truth masks}, but yielding higher-resolution predictions, namely $240\times320$ pixels,
	\item {\it \textbf{coarse}} segmentation model: trained with both NIR and VIS data from the v1 and v2 versions of the Warsaw-Biobase-Postmortem-Iris database for 120 epochs, but this time with \textbf{coarse ground truth masks}, denoting only the inner and the outer iris boundary and eyelids, also producing masks in $240\times320$ size.
\end{enumerate} 

\begin{figure}[h!]
\centering
{\it \textbf{fine}} and {\it \textbf{fine v2highres}} \textbf{CNN} segmented images, normalized images, and normalized masks\\\vskip1mm
\includegraphics[width=0.4\linewidth]{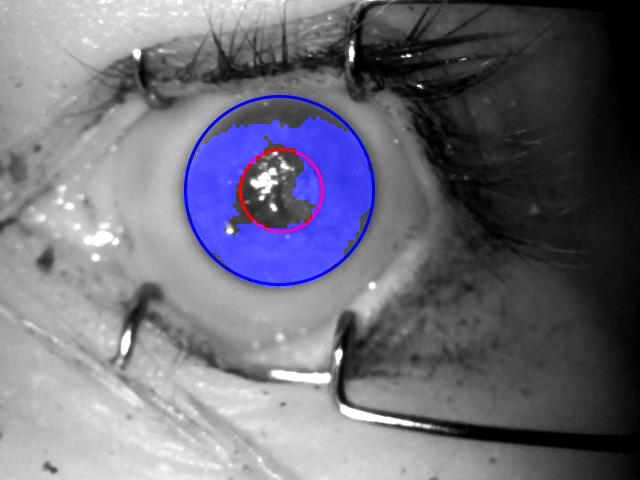}\hskip1mm
\includegraphics[width=0.4\linewidth]{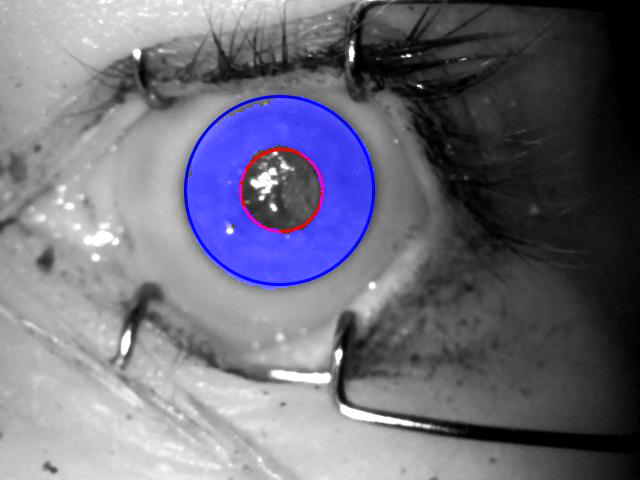}\\\vskip1mm
\includegraphics[width=0.4\linewidth]{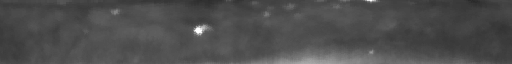}\hskip1mm
\includegraphics[width=0.4\linewidth]{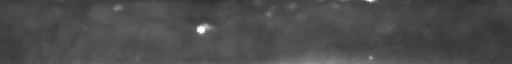}\\\vskip1mm
\includegraphics[width=0.4\linewidth]{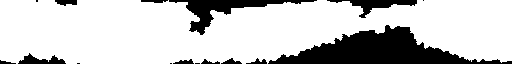}\hskip1mm
\includegraphics[width=0.4\linewidth]{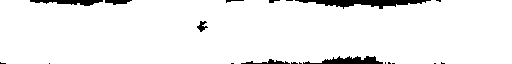}
\caption{Segmentation results for post-mortem sample from Fig. \ref{fig:models:examples}, obtained from the {\it \textbf{fine}} and {\it \textbf{fine v2highres}} segmentation models.}
\label{fig:models_fine:examples}
\end{figure}

\subsection{Iris and mask normalization}
For the proposed method to serve as a drop-in replacement for the Daugman method recognition pipeline, the prediction obtained from the DCNN-based segmentation has to be correctly normalized onto a dimensionless polar-coordinate rectangle. For this stage, a method for localizing pupillary and limbic iris boundaries has been devised, which employs a Hough transform that is applied to the prediction obtained from the {\it \textbf{coarse}} segmentation model, {similarly to the methodology introduced in \cite{KerriganICB2019Segmentation} and \cite{HOFBAUER201917}}, as this model yields the smoothest prediction. These boundary parameters are then used in all subsequent experiments, including those involving the \emph{fine} and \emph{fine v2highres} models. The segmentation methodology employing the \emph{coarse} segmentation model is visualized in Fig. \ref{fig:models:examples}. Similar results are then presented in Fig. \ref{fig:models_fine:examples} for the two remaining fine-grained models. Notably, the original \emph{fine} model from \cite{TrokielewiczIWBF2018} seems to over-aggressively mask out some portions of the iris, while at the same time mistakenly denoting some portions of the pupil as iris. This is fixed in the \emph{fine v2highres} model, which was trained with twice the iterations count, cf. right side of Fig. \ref{fig:models_fine:examples}.

\subsection{IriCore: a commercial benchmark method}
In addition to evaluating our approach against the unmodified OSIRIS algorithm, we also choose to compare it with one of the leading iris matchers from commercial vendors, namely the IriCore \cite{IriCore}, which offered the best performance out of four matchers in our previous evaluations of post-mortem decay impact on various iris recognition methodologies \cite{TrokielewiczTIFS2018}. {This commercial matcher was also ranked as one of the two best matchers in NIST IREX I, it has the STQC certification and is part of the world largest biometric project in India (AADHAAR).} IriCore comes from the same manufacturer as the IriShield camera used to collect post-mortem samples in our studies -- and therefore it may have an additional advantage in this particular task. Since IriCore follows an undisclosed iris recognition algorithm, and can be only used as a black box taking two images and producing a dissimilarity score between them. The distributions of these dissimilarity scores, which are expected to be ranging from 0 to 1.1 for genuine comparisons and from 1.1 to 2.0 for impostor ones, are then used to construct ROC curves which can be compared against those obtained from original and modified OSIRIS. No additional score normalization is introduced for IriCore.

\section{Iris recognition accuracy evaluation}
\label{sec:experiments}
\subsection{Comparison score generation}
In this Section we evaluate the new segmentation approach introduced in Sec. \ref{sec:Methods} by injecting the segmentation results obtained from the three DCNN-based models into the OSIRIS recognition pipeline, and also comparing their performance with unmodified OSIRIS algorithm, as well as with the accuracy offered by the IriCore. The evaluation is performed on two test databases described in Sec. \ref{sec:Database}. Generation of genuine and impostor comparison score distributions was done by performing comparisons between all possible iris image pairs. One comparison per a given image pair was performed, \ie if image A was matched against image B, then image B is not matched against image A.

For the post-mortem data, eight separate score distributions are generated for both the genuine and impostor comparisons, taking into account samples collected during different, progressing observation horizons in respect to time that has elapsed since a subject's death. These include scores obtained from samples collected fewer than 10 hours after death, fewer than 24 hours, 48 hours, 60 hours, 110 hours, 160 hours, 210 hours, and finally 370 hours (all available samples). This is to evaluate the recognition accuracy in respect to the increasing post-mortem interval. All possible comparison pairs were evaluated here as well.

\begin{table}[!ht]
\renewcommand{\arraystretch}{1.2}
\caption{{Summary of the Equal Error Rates (EERs) obtained when performing all possible genuine and impostor comparisons in data subsets as defined in Sec. \ref{sec:experiments} for the two benchmark methods and the best proposed method ({\bf fine v2highres} model). Numbers of comparisons for each experiment are included in the last column.}}
\label{table:eers}
\centering\footnotesize
\begin{tabular}[t]{|c|c|c|c|c|}
\hline
{\bf Data subset} & {\bf original OSIRIS} & {\bf IriCore} & {\bf proposed} & number of comparisons \\
 & [\%] &[\%]& [\%] & \\
\hline
\hline
{$\leq$ 10h postmortem} & 16.89  & 5.37  & \cellgreen{\bf 0.96} & 46 360\\
\hline
{$\leq$ 24h postmortem} & 18.73  & 6.31  & \cellgreen{\bf 1.36} & 61 425 \\
\hline
{$\leq$ 48h postmortem} & 20.34  & 8.00  & \cellgreen{\bf 3.38} & 88 831 \\
\hline
{$\leq$ 60h postmortem} &  23.69 & 10.46 & \cellgreen{\bf 7.18} & 178 503 \\
\hline
{$\leq$ 110h postmortem} & 24.72 & 14.36 & \cellgreen{\bf 10.36} & 296 065\\
\hline
{$\leq$ 160h postmortem} & 28.68 & 17.79 & \cellgreen{\bf 15.02} & 405 450 \\
\hline
{$\leq$ 210h postmortem} & 29.94 & 20.77 & \cellgreen{\bf 17.60} & 476 776 \\
\hline
{$\leq$ 369h postmortem} & 33.59 & 25.38 & \cellgreen{\bf 21.95} & 597 871 \\
\hline
\hline
{disease data} & 8.90 & 3.97  & \cellgreen{\bf 2.55} & 152 076 \\
\hline

\end{tabular}
\end{table}

\subsection{Recognition accuracy: post-mortem dataset}
The results are presented in the form of receiver operating characteristic (ROC) curves, which plot the correct match rates against false match rates. In addition, equal error rates (EER) are calculated and denoted in the plot legend. Fig. \ref{fig:ROCs:cold_short} shows ROC curves denoting the performance of the baseline methods (OSIRIS and IriCore) performance compared against performance obtained by employing image segmentation results from the three DCNN-based models for the post-mortem data in four shortest post-mortem time horizons, comprising samples collected earlier than 60 hours after the demise of each subject. Raw Hamming distance scores are plotted in solid lines, whereas the scores after score normalization (cf. Sec. \ref{sec:OSIRISintro}) are plotted with dotted lines. Fig. \ref{fig:ROCs:cold_long} contains the analogous graphs, but for the remaining four longer time horizons, namely those with samples collected up to 369 hours post-mortem. {Equal Error Rates for all observation horizons for the two benchmark methods, as well as the proposed approach (winning {\bf fine v2highres} model), are summarized in Tab. \ref{table:eers}.}

Notably, very good performance can be achieved using each one of the proposed models for samples collected up to 10 hours after death, with EER as low as 0.96\%, which only slightly increases to 1.36\% when samples collected up to 24 hours post-mortem are added to the database. For samples collected up to 48 hours (two days) after death, the lowest EER offered by the \emph{fine v2highres} model is 3.38\%, which means that the method can still be usable. As more difficult samples are being added to the database, the errors increase, reaching 21.95\% for the best performing \emph{fine v2highres} model. Although the \emph{fine} model given better EER, it also presents a risk of higher false match rates than the \emph{coarse} model (see the region of the ROC curve near the Y axis). This can be alleviated by introducing score normalization (cf. Sec. \ref{sec:OSIRISintro}), but this in turn increases the EER by almost 3\%. The \emph{fine v2highres} model, on the other hand, seems to offer similar risk of false matches as the \emph{fine} model with score normalization, but at the same time preserving most of the overall performance with EER only slightly higher than this of the \emph{fine} model. Original OSIRIS, on the other hand, is incomparable to the proposed approach from the very beginning, yielding EER of almost 17\% for the `easiest' samples, and the EERs from this method are increasing steadily up to 33.59\% for the entire database (samples collected up to 370 hours post-mortem). The IriCore's performance starts with EER=5.37\% for the easiest samples, which is still more than 4 percentage points behind our proposed solution, and decreases steadily with the increasing post-mortem time horizon, reaching 25.38\% for the most challenging set of samples (all images). The only advantage the IrisCore may have over our approach is better performance in selected time post-mortem horizons (samples collected up to 60, 110 and 160 hours) in very low FMR registers.

\subsection{Recognition accuracy: elderly/ophthalmic conditions dataset}
\begin{figure}[t]
\centering
\includegraphics[width=0.45\linewidth]{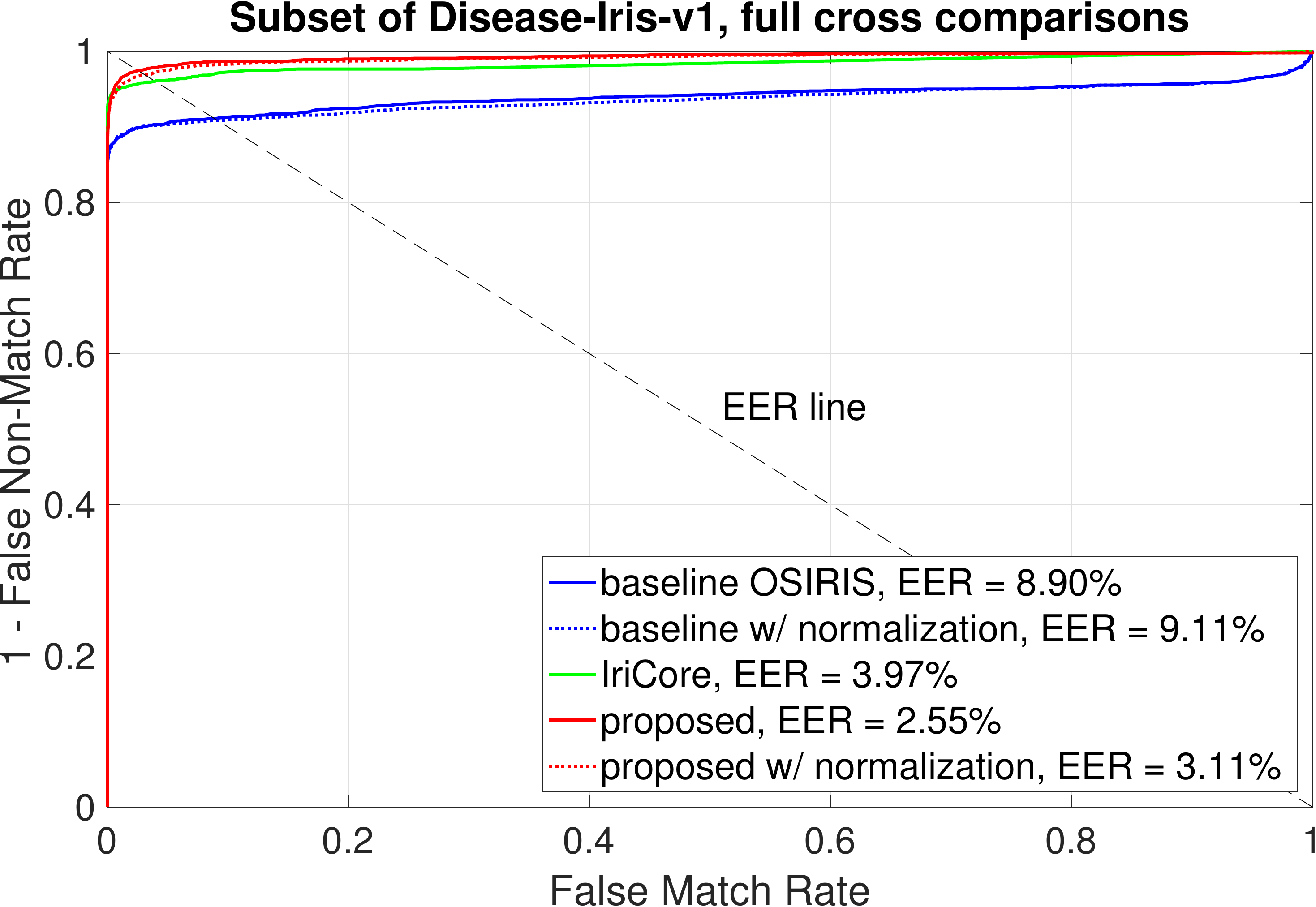}\hskip1mm
\includegraphics[width=0.53\linewidth]{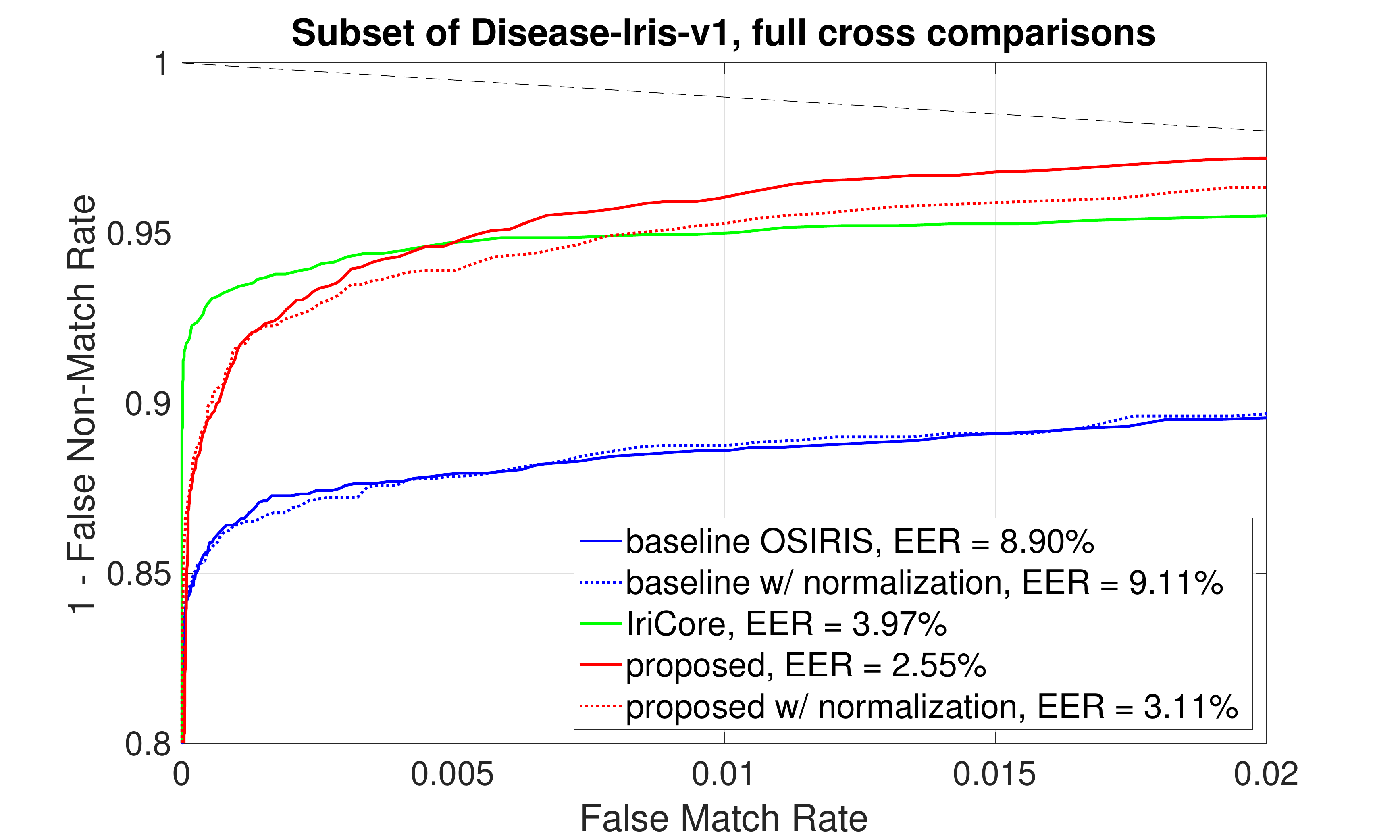}
\caption{ROC curves obtained for the Warsaw-BioBase-Disease-Iris-v1 dataset. Original (solid lines) and normalized (dotted lines) scores are shown. The plot on the right is a close-up of the one on the left.}
\label{fig:ROCs:disease}
\end{figure}

Fig. \ref{fig:ROCs:disease} presents ROC curves obtained when matching samples from the Warsaw-BioBase-Disease-Iris-v1 databases using a baseline OSIRIS approach and the proposed segmentation approach based on the \emph{coarse} model. Because this dataset does not contain any post-mortem samples, there is no justification to apply post-mortem-aware fine-grained segmentation model here. 

Because this test database contains images collected from elderly people, often with various ophthalmic conditions -- also challenging ones, the advantage of the proposed approach is evident, as it offers EER=2.55\% compared to 8.9\% yielded by OSIRIS and 3.97\% obtained from IriCore.

\subsection{Discussion of score normalization}
\begin{figure}[t]
\centering
\includegraphics[width=0.49\linewidth]{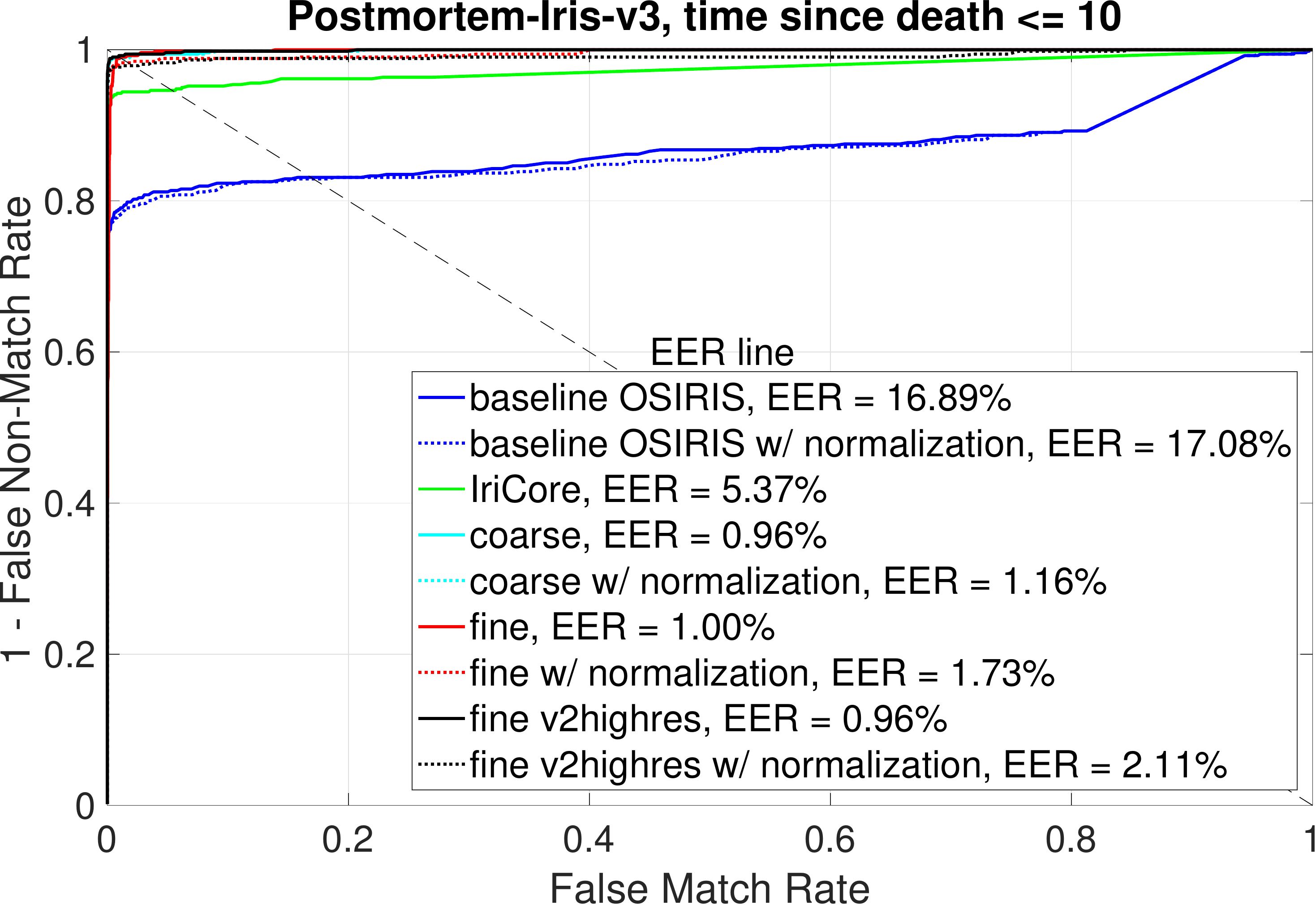}\hskip1mm
\includegraphics[width=0.5\linewidth]{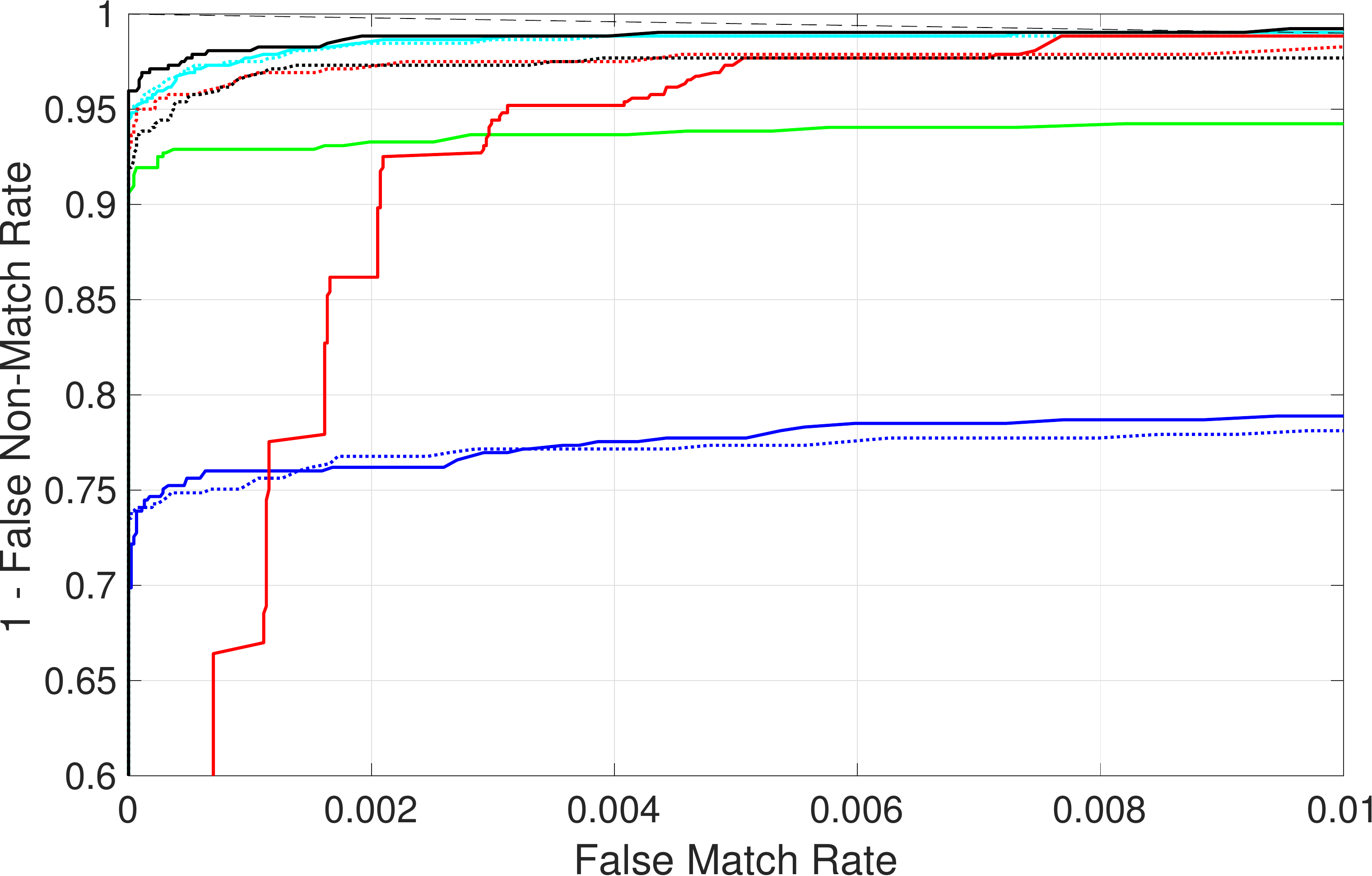}\\\vskip2mm
\includegraphics[width=0.49\linewidth]{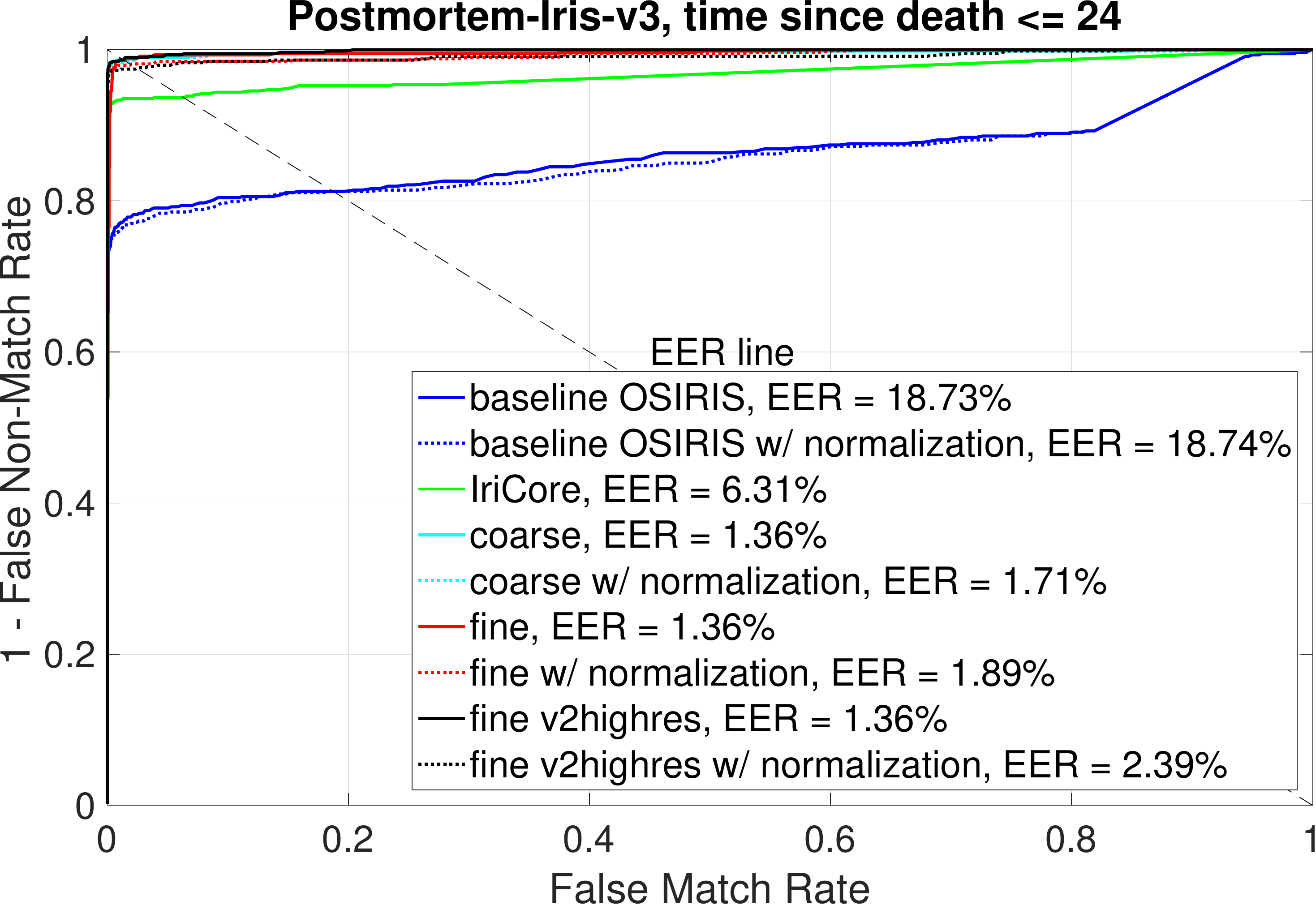}\hskip1mm
\includegraphics[width=0.5\linewidth]{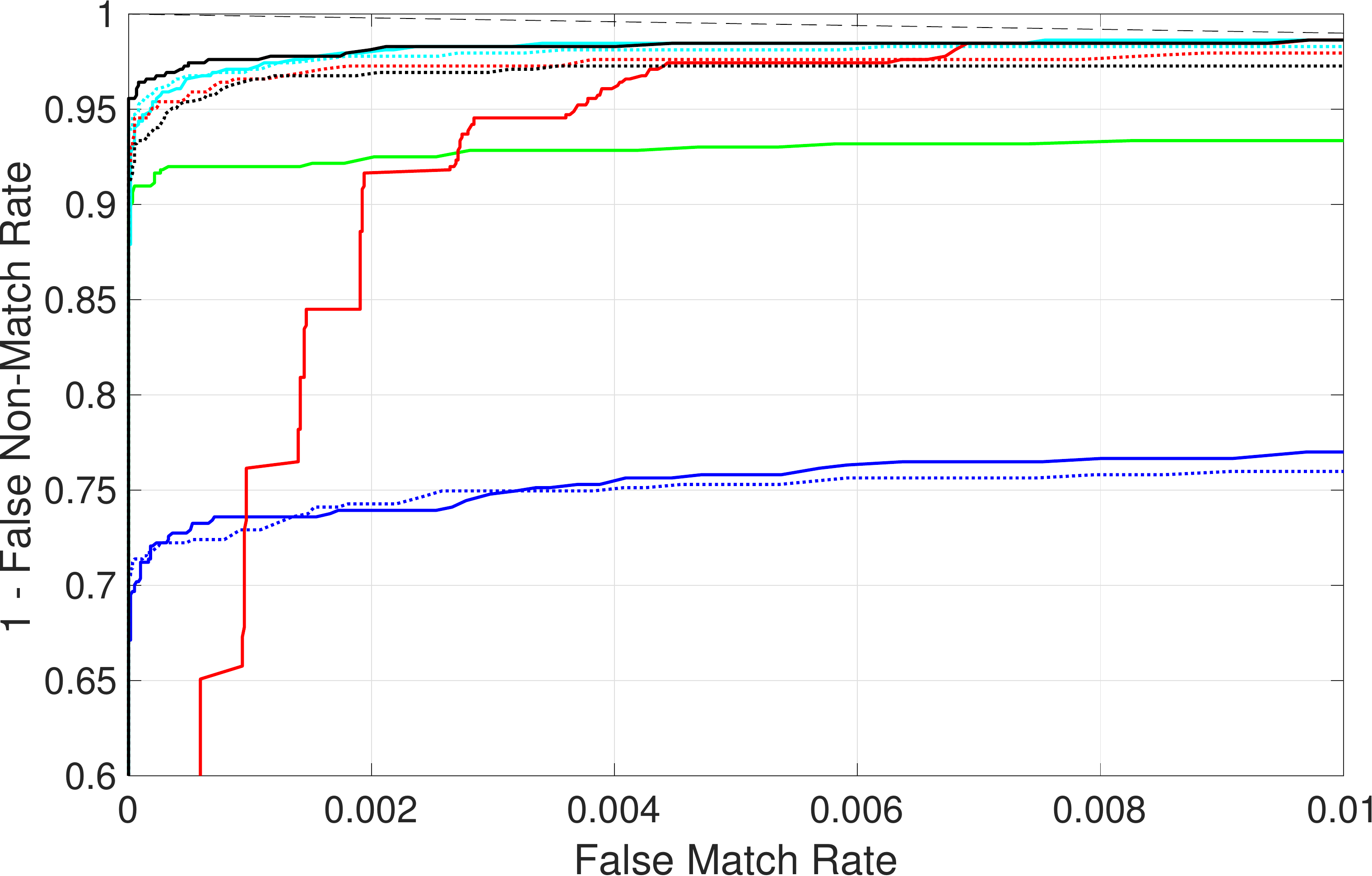}\\\vskip2mm
\includegraphics[width=0.49\linewidth]{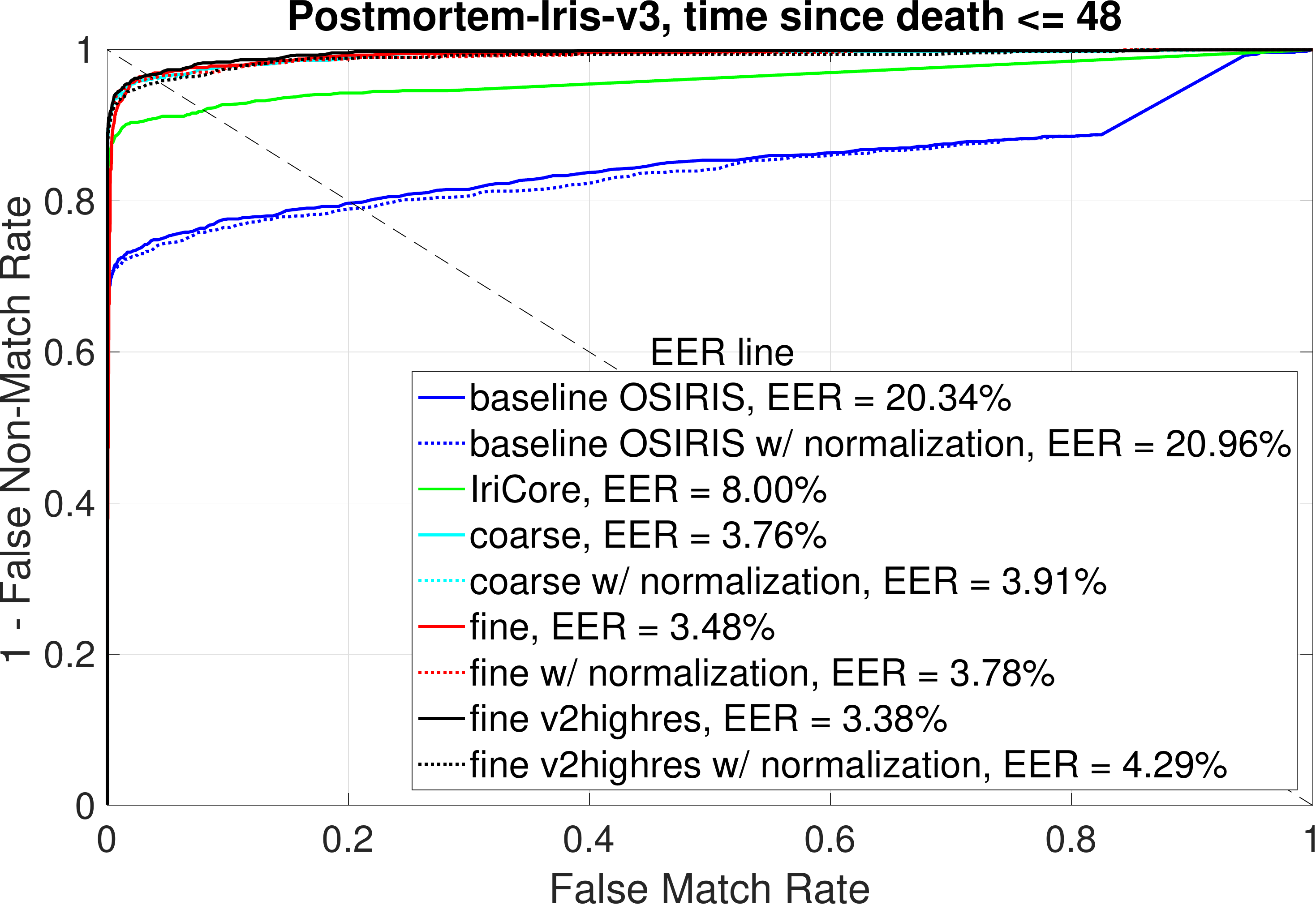}\hskip1mm
\includegraphics[width=0.5\linewidth]{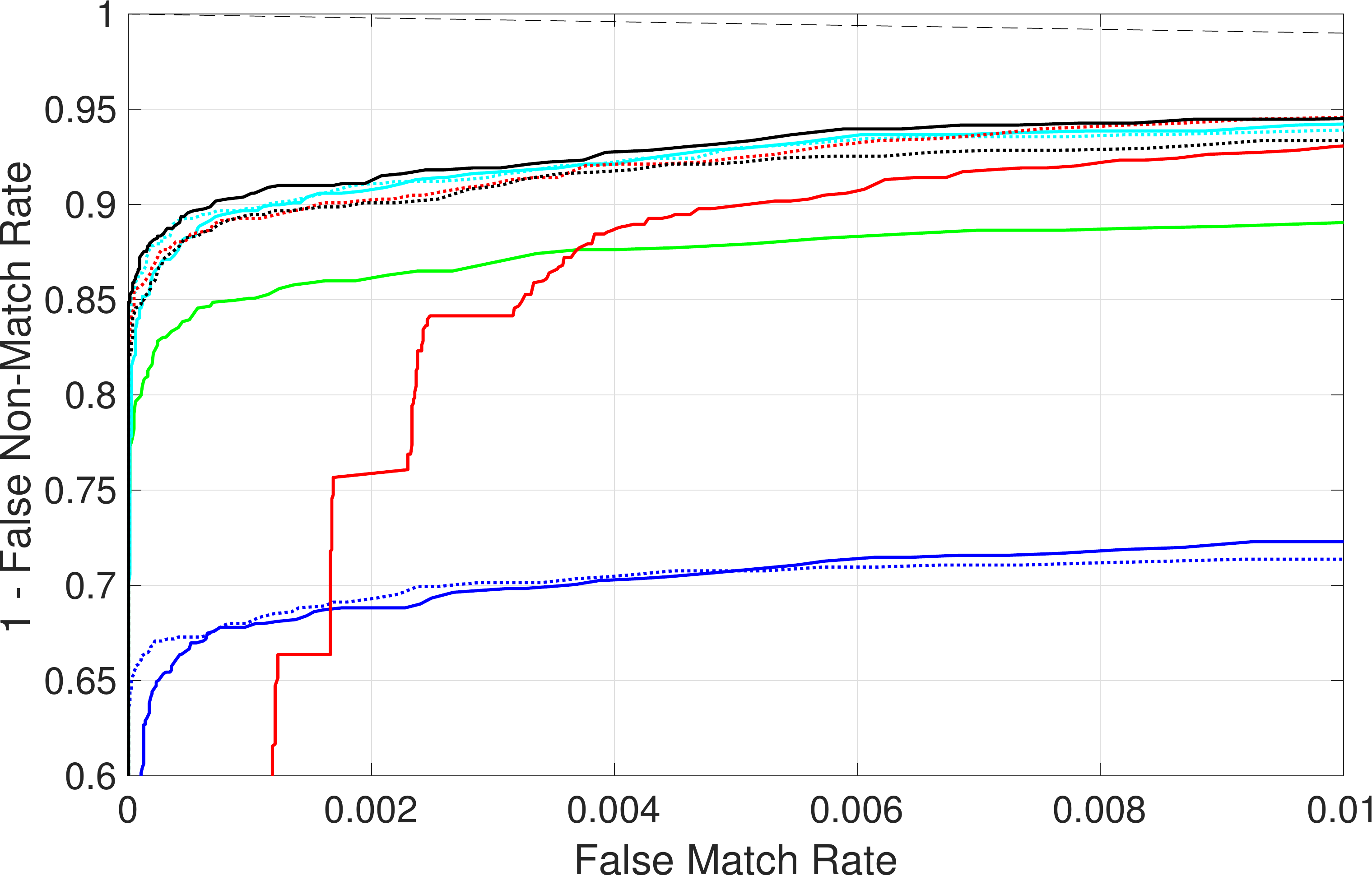}\\\vskip2mm
\includegraphics[width=0.49\linewidth]{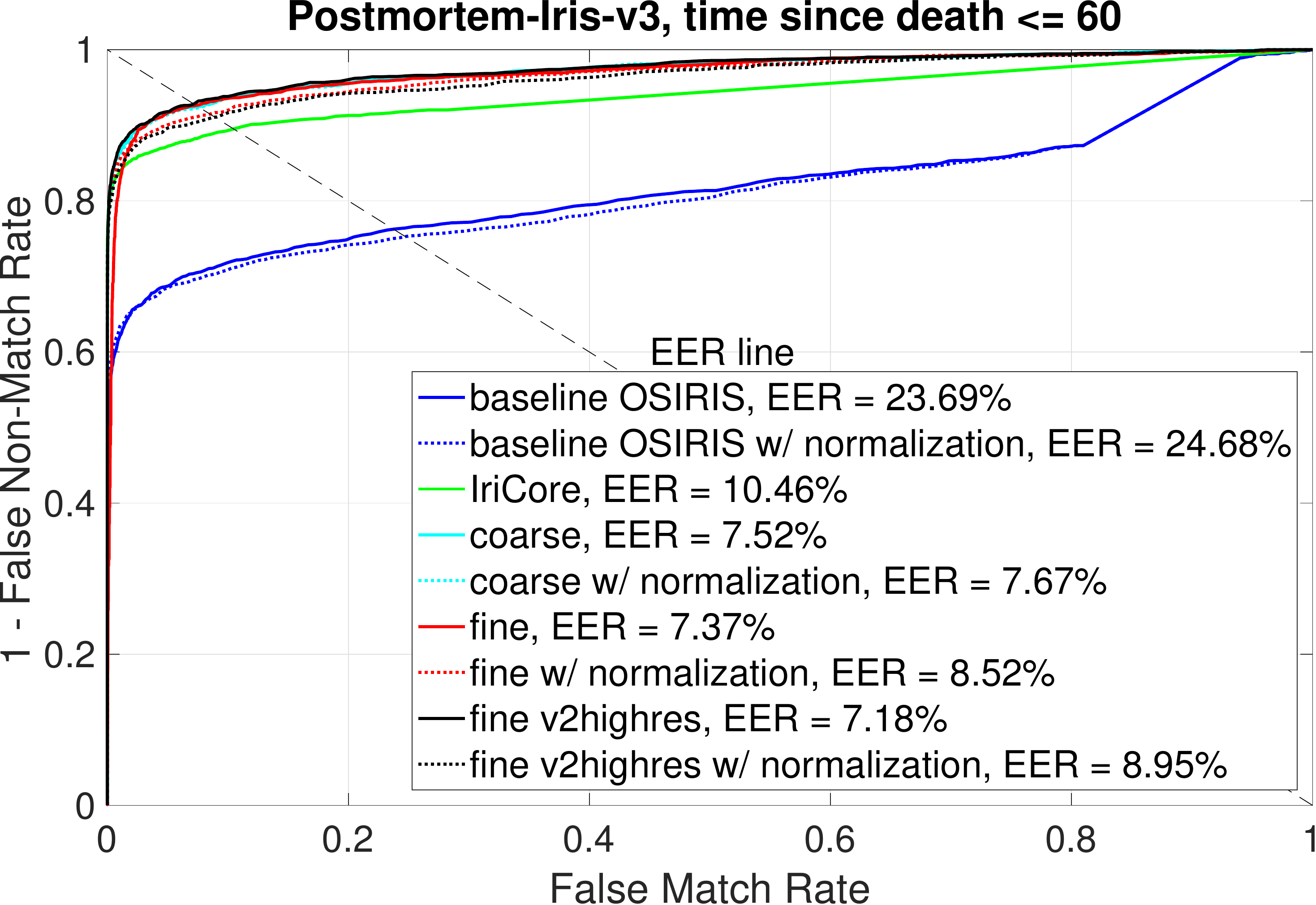}\hskip1mm
\includegraphics[width=0.5\linewidth]{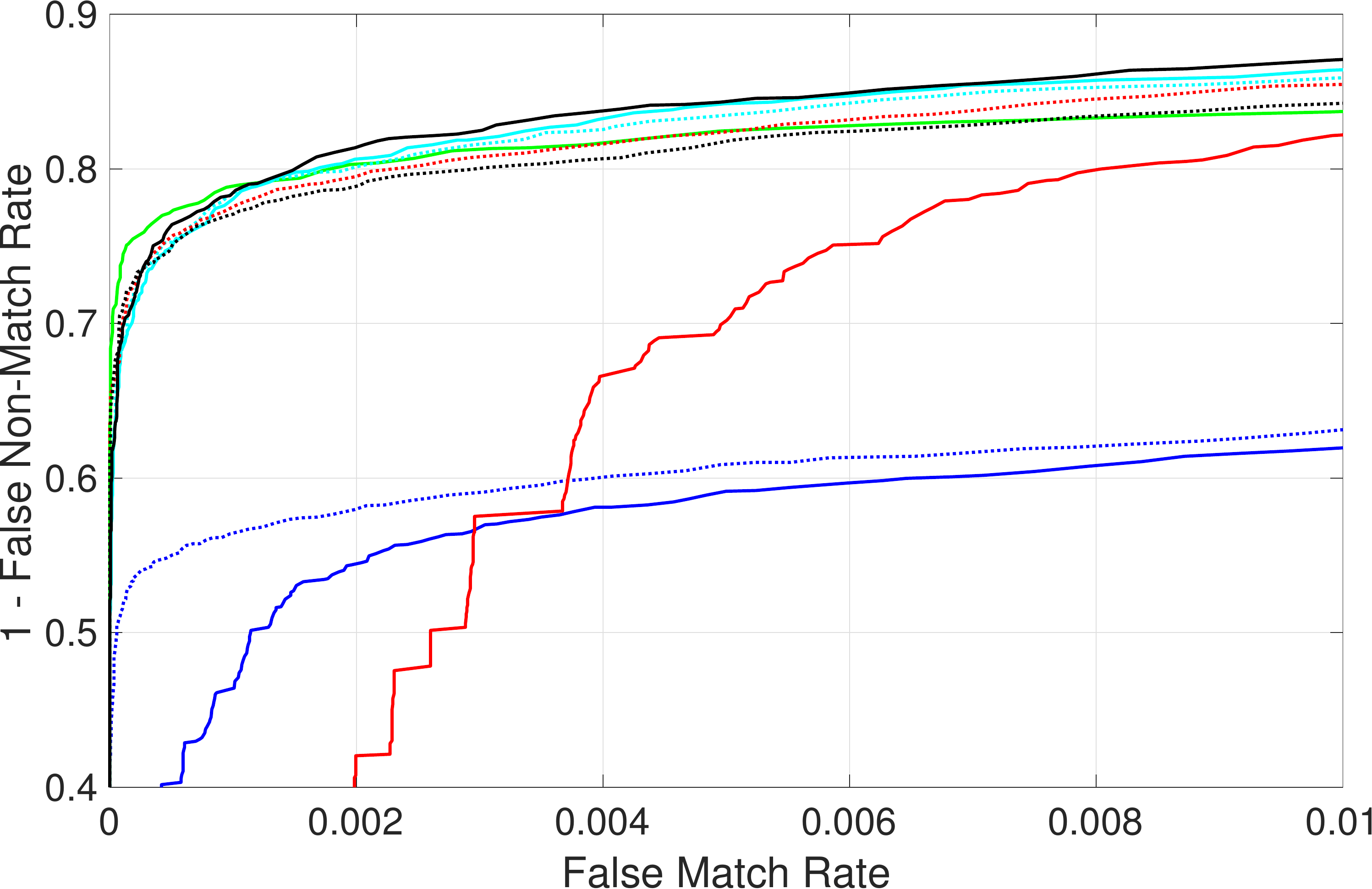}
\caption{Receiver Operating Characteristic curves obtained when comparing post-mortem samples with different observation time horizons. Results for the baseline OSIRIS performance, IriCore benchmark method, and three modifications introduced by the Authors are presented. Original (solid lines) and normalized (dotted lines) scores are shown. Plots on the right are close-ups of the ones on the left in each pair.}
\label{fig:ROCs:cold_short}
\end{figure}

\begin{figure}[t]
\centering
\includegraphics[width=0.49\linewidth]{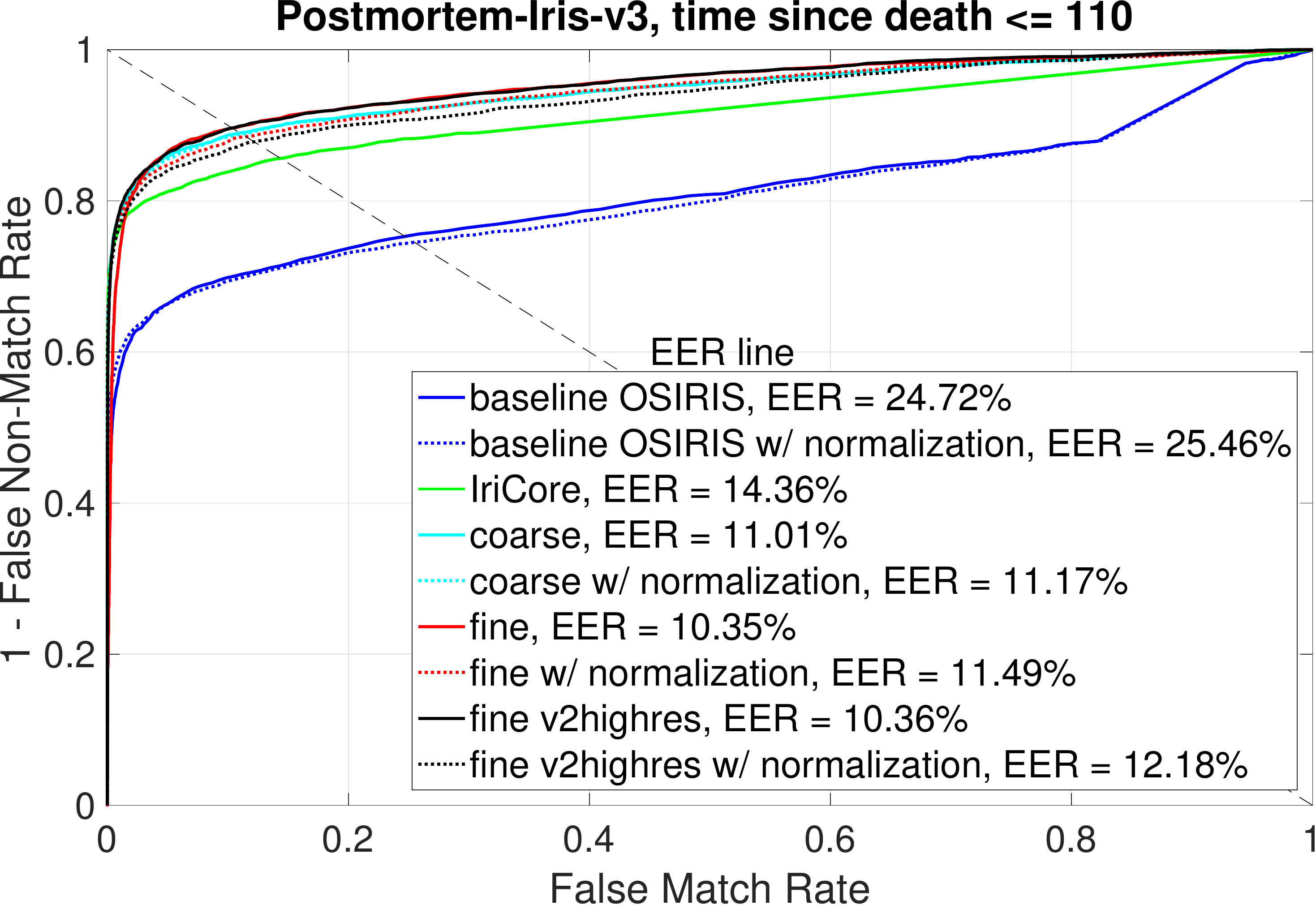}\hskip1mm
\includegraphics[width=0.5\linewidth]{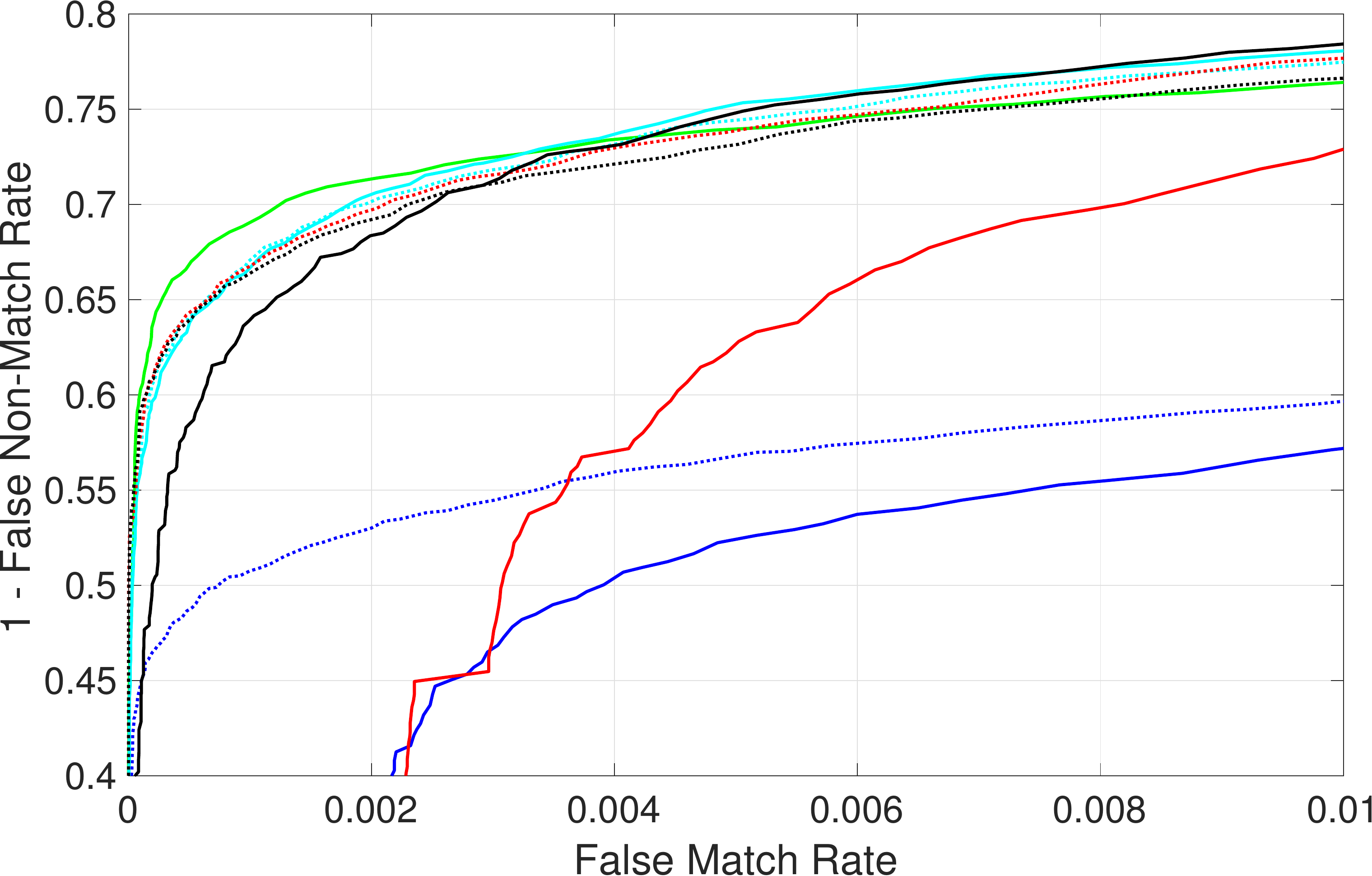}\\\vskip2mm
\includegraphics[width=0.49\linewidth]{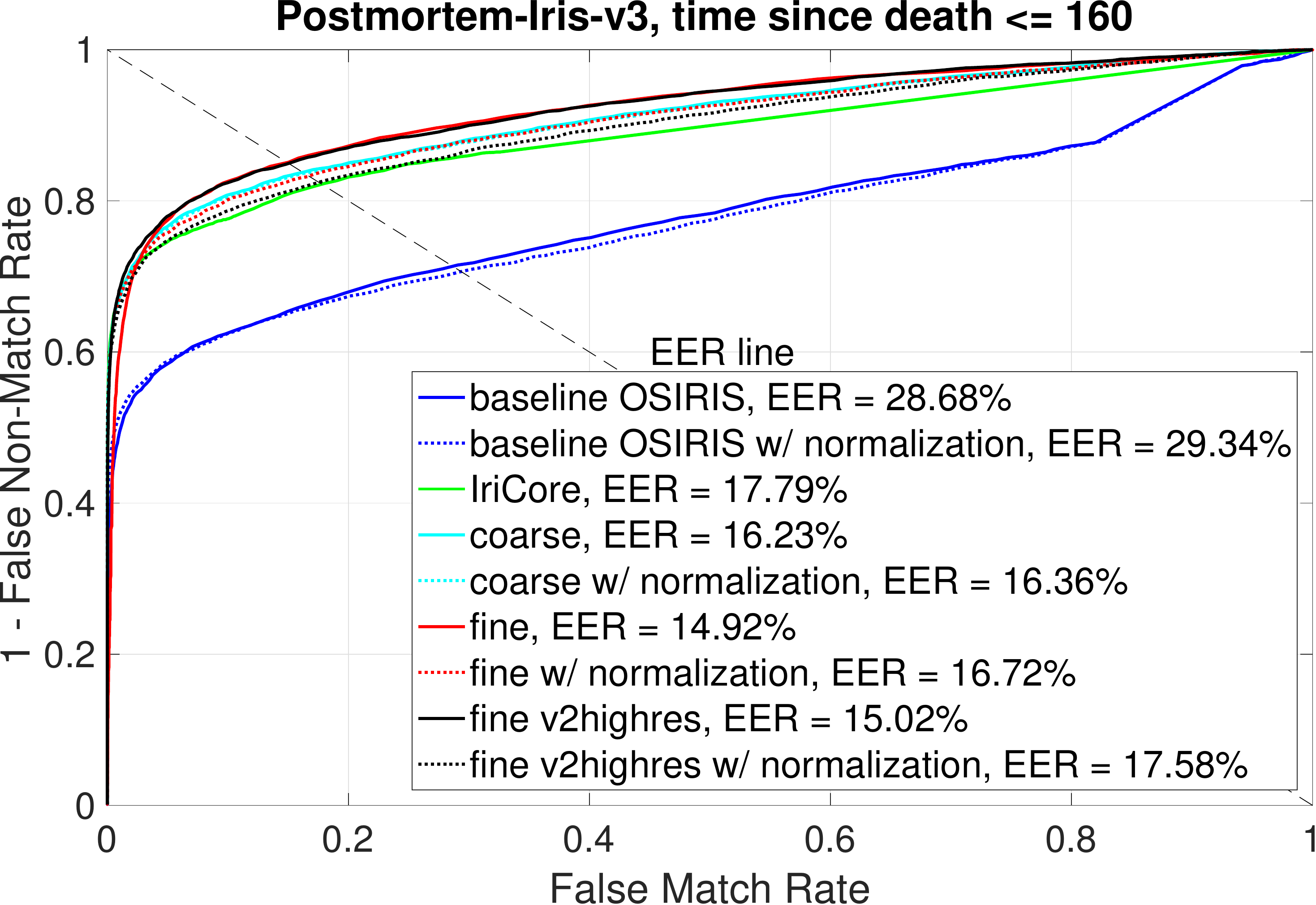}\hskip1mm
\includegraphics[width=0.5\linewidth]{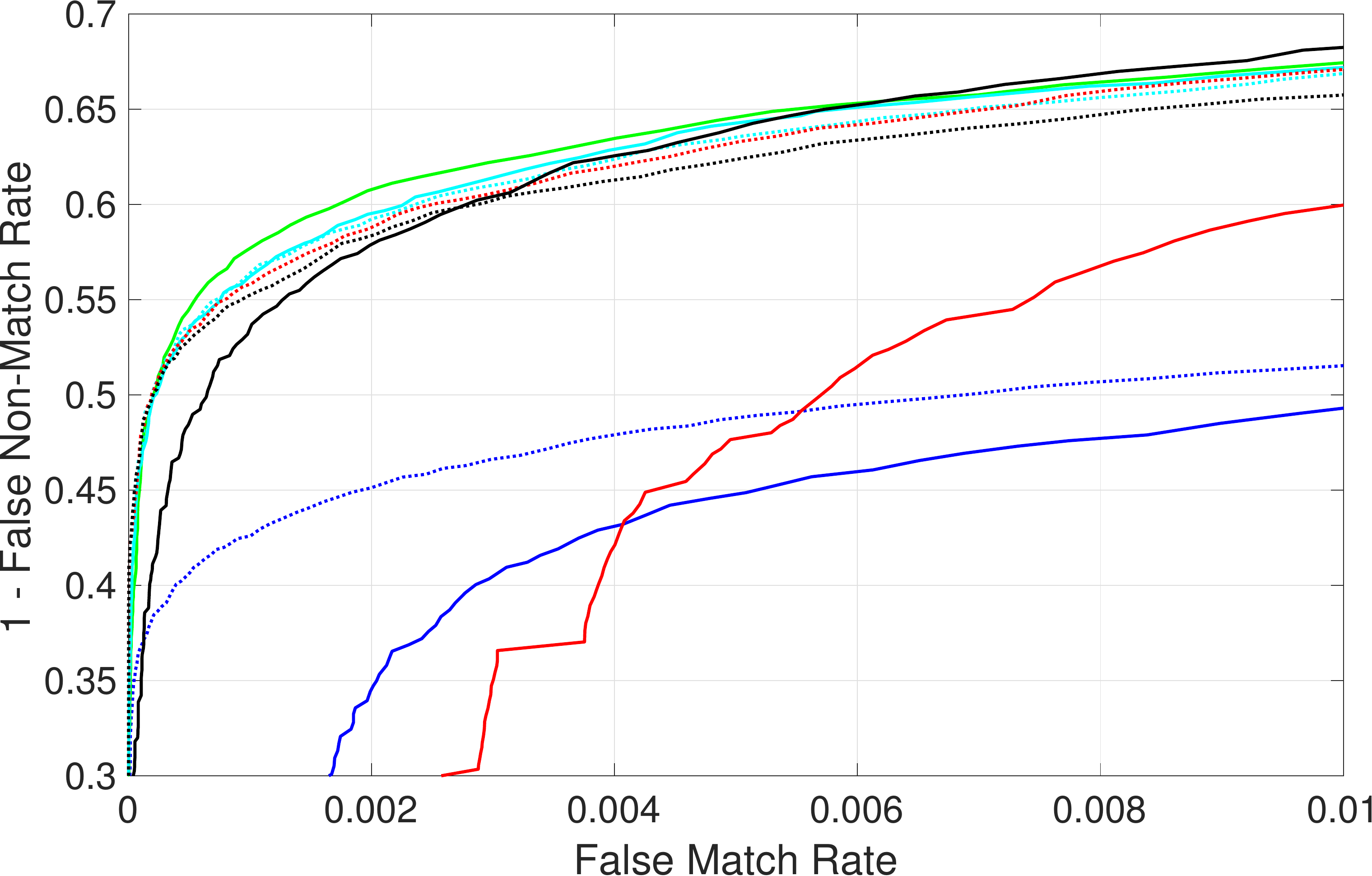}\\\vskip2mm
\includegraphics[width=0.49\linewidth]{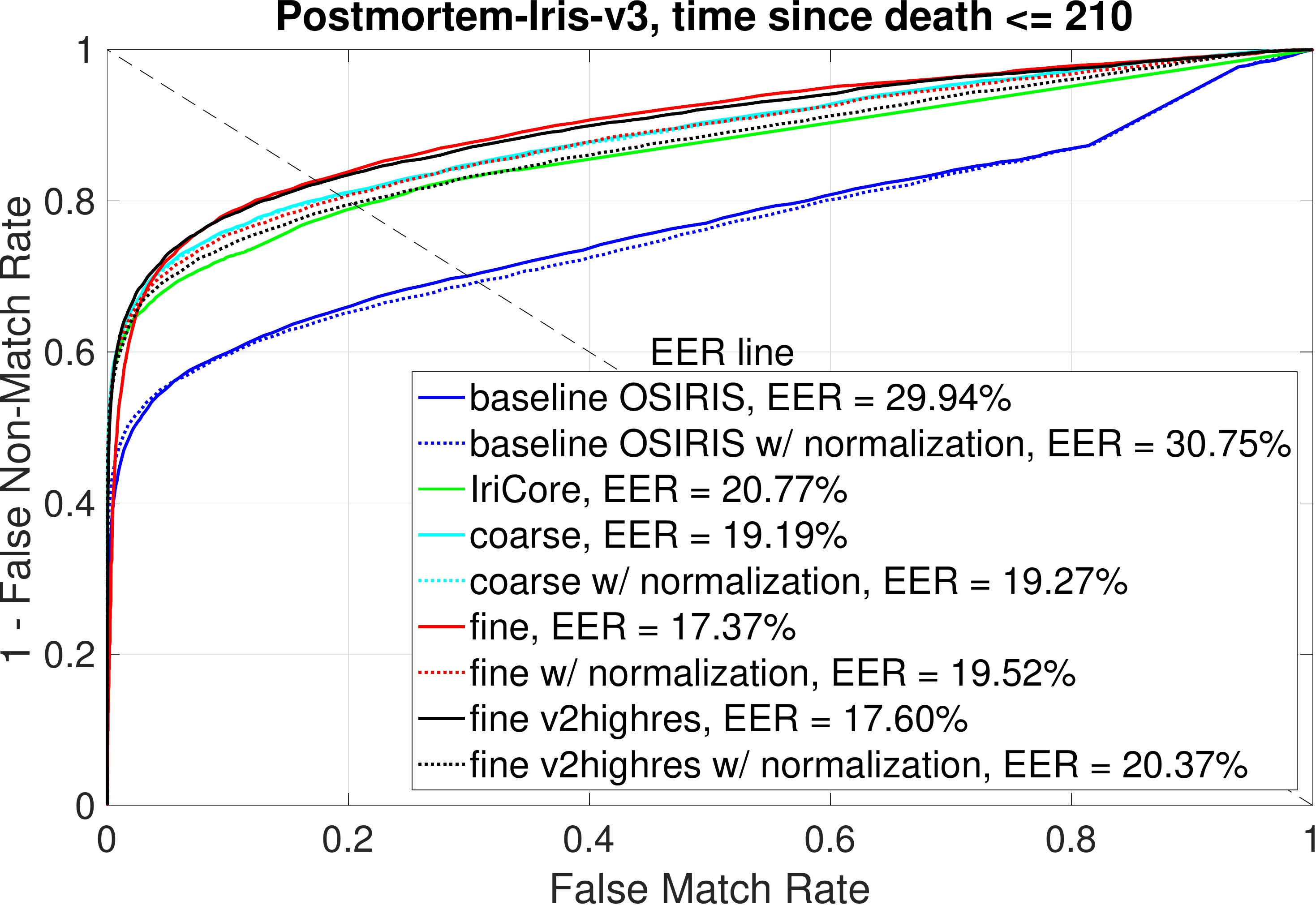}\hskip1mm
\includegraphics[width=0.5\linewidth]{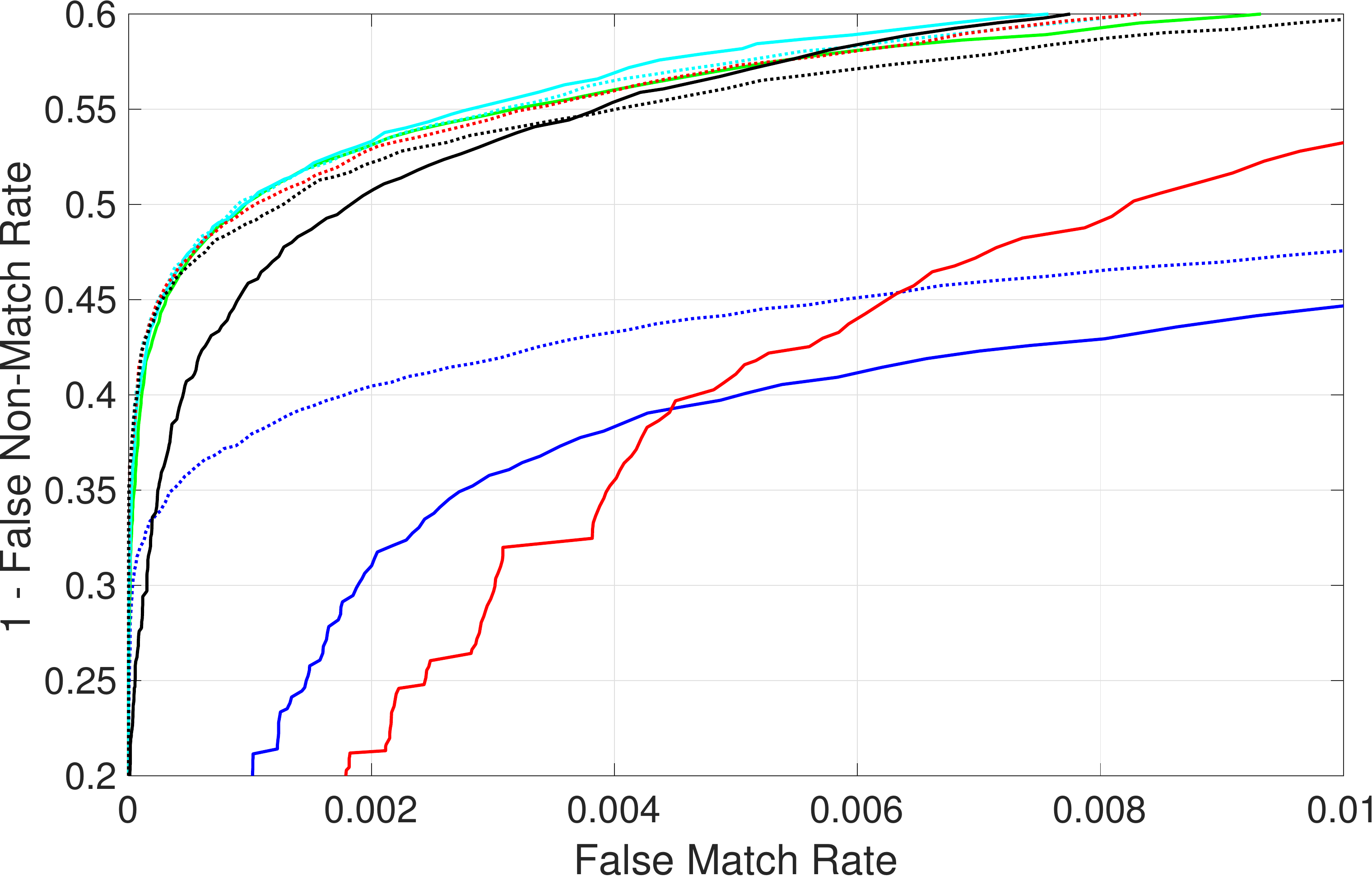}\\\vskip2mm
\includegraphics[width=0.49\linewidth]{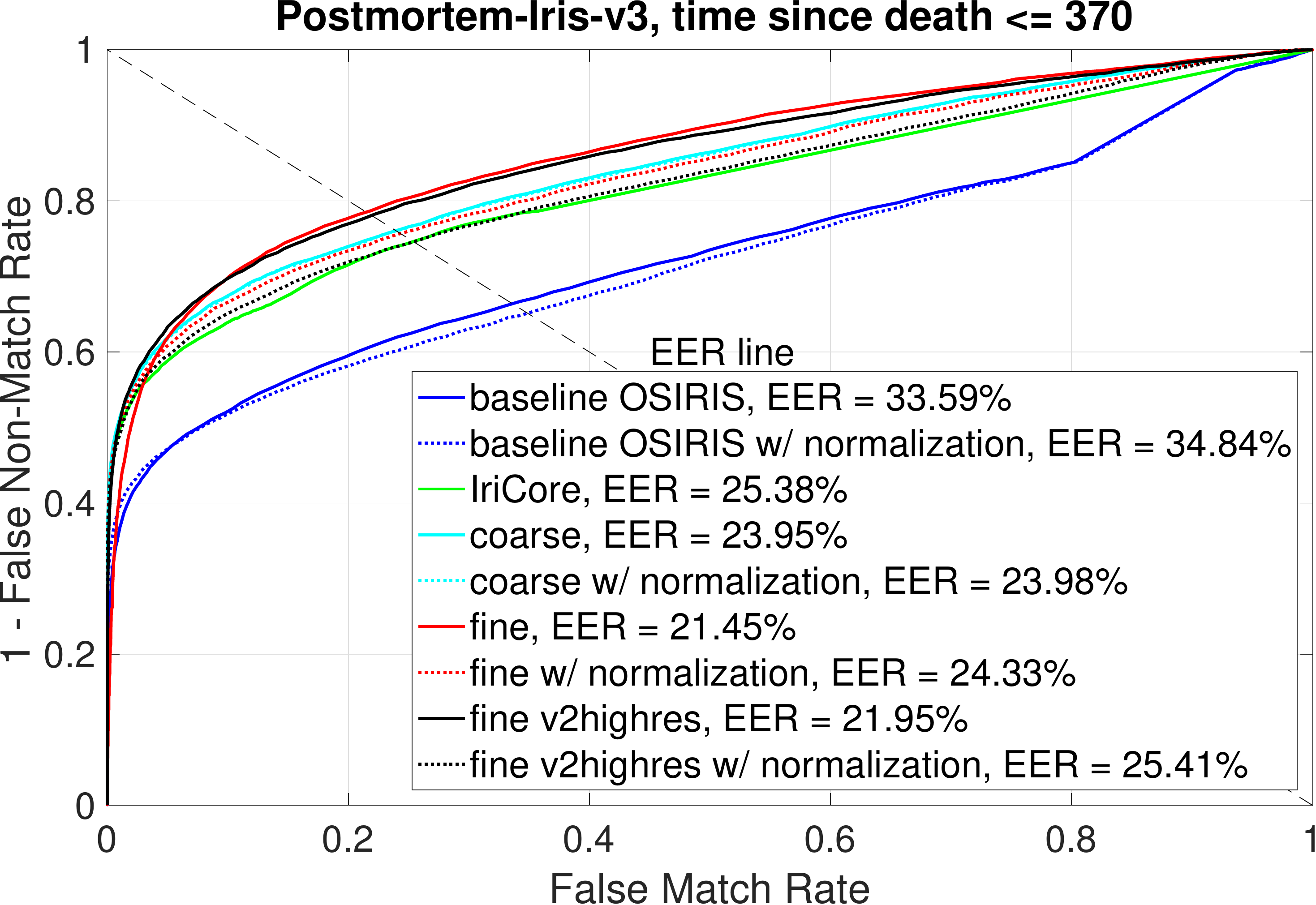}\hskip1mm
\includegraphics[width=0.5\linewidth]{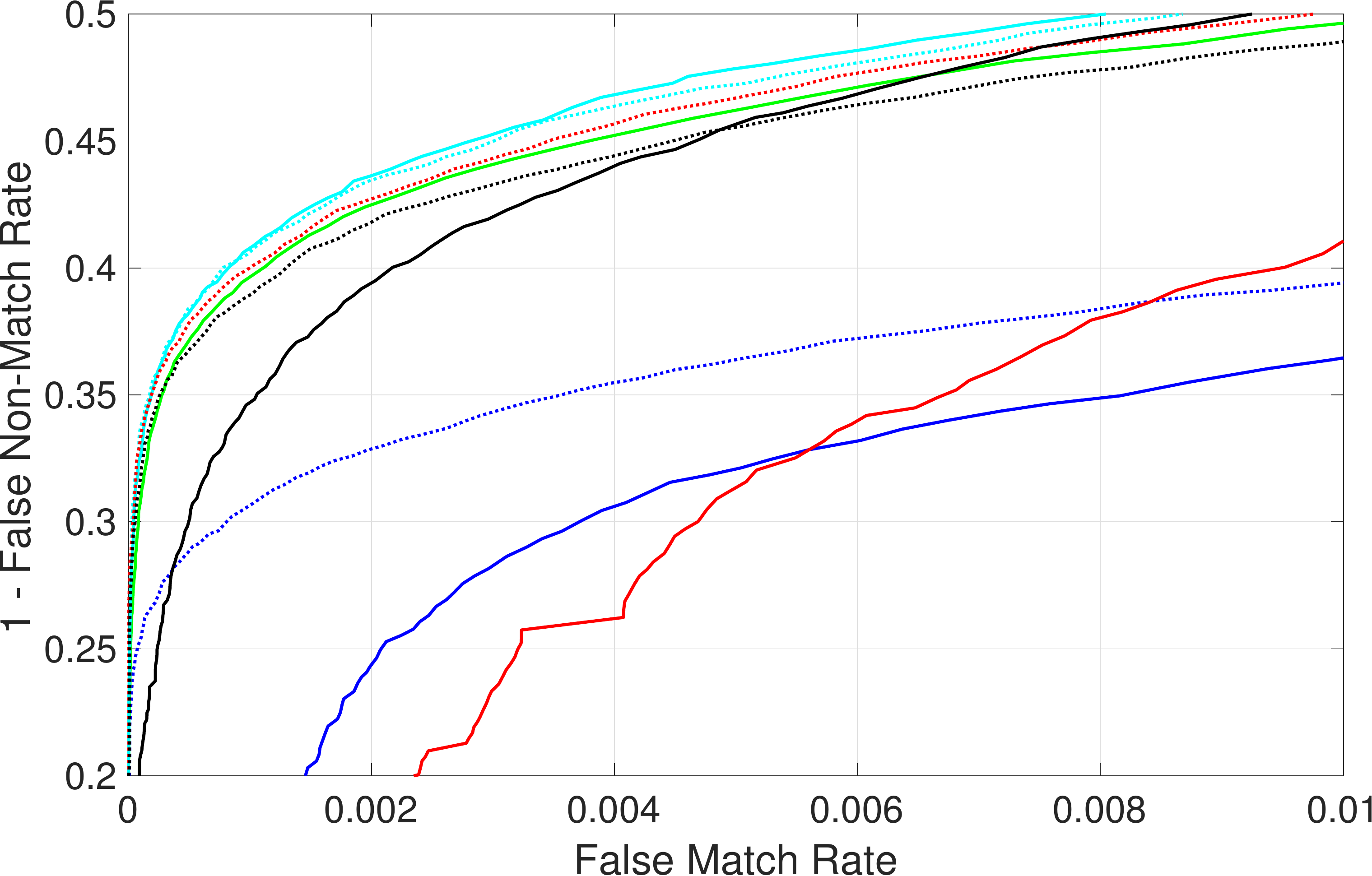}
\caption{Same as in Fig. \ref{fig:ROCs:cold_short}, but plotted for comparison scores obtained from long-term samples collected from 210 up to 369 hours post-mortem.}
\label{fig:ROCs:cold_long}
\end{figure}

As argued in Sec. \ref{sec:OSIRISintro}, score normalization can help rectify significant portions of false matches that were caused by comparing small numbers of commonly unmasked bits in two iris codes. As seen in Fig. \ref{fig:ROCs:cold_short}, employing this normalization for the very selective segmentation model, namely the old \emph{fine} model (red plots) can vastly reduce the false matches at a little EER-related cost. With less selective segmentation, however, the normalization cannot offer any advantage over lack of such, \eg in the best performing \emph{fine v2 highres model} (black plots). With increasing time horizons, as in Fig. \ref{fig:ROCs:cold_long}, even the less selective models start to mask out iris portions more aggressively, hence the score normalization can improve the performance in low FMR registers of most of the models.

\subsection{{Rank list performance}}
\begin{figure}[h!]
\centering
\includegraphics[width=0.49\linewidth]{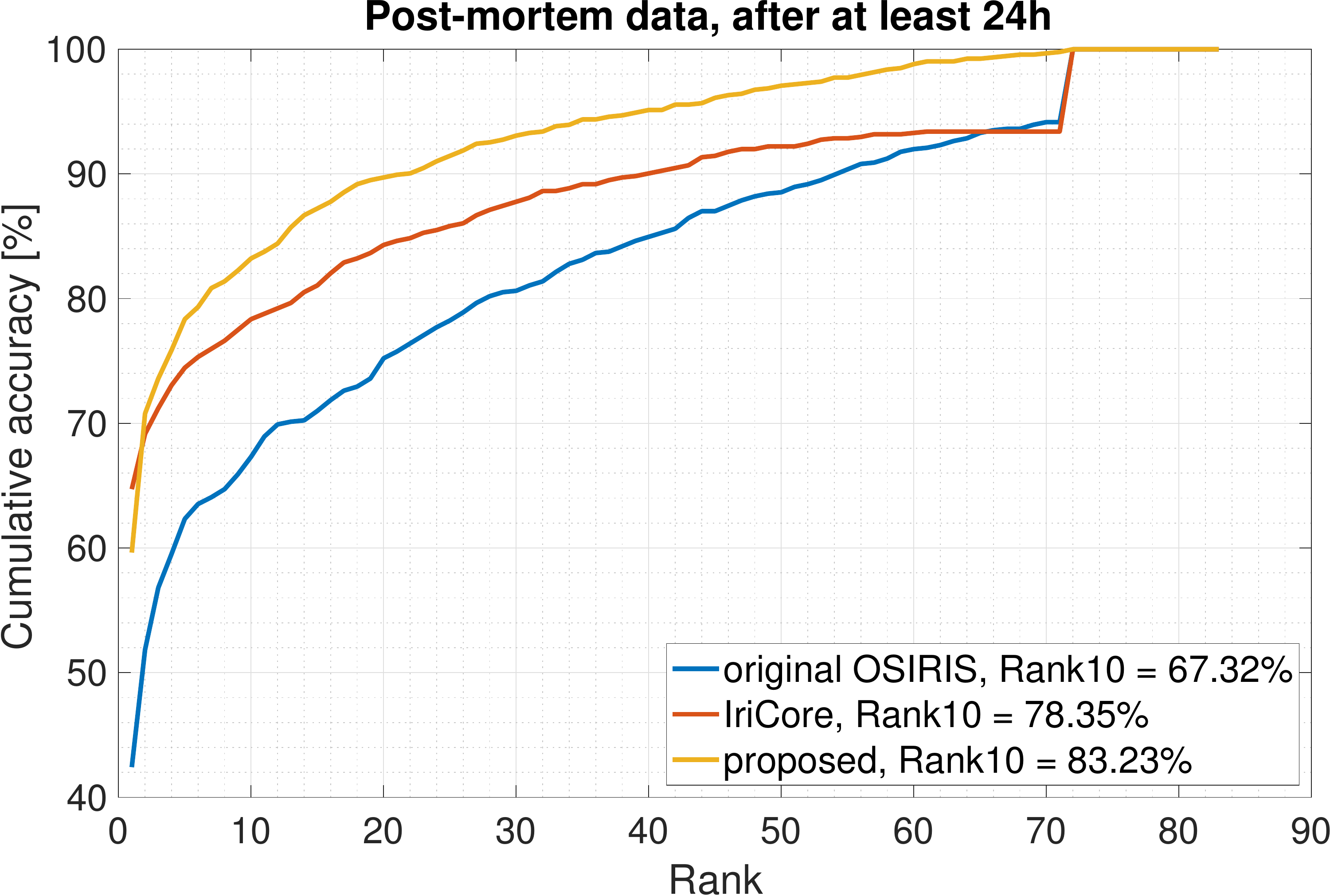}\hskip1mm
\includegraphics[width=0.49\linewidth]{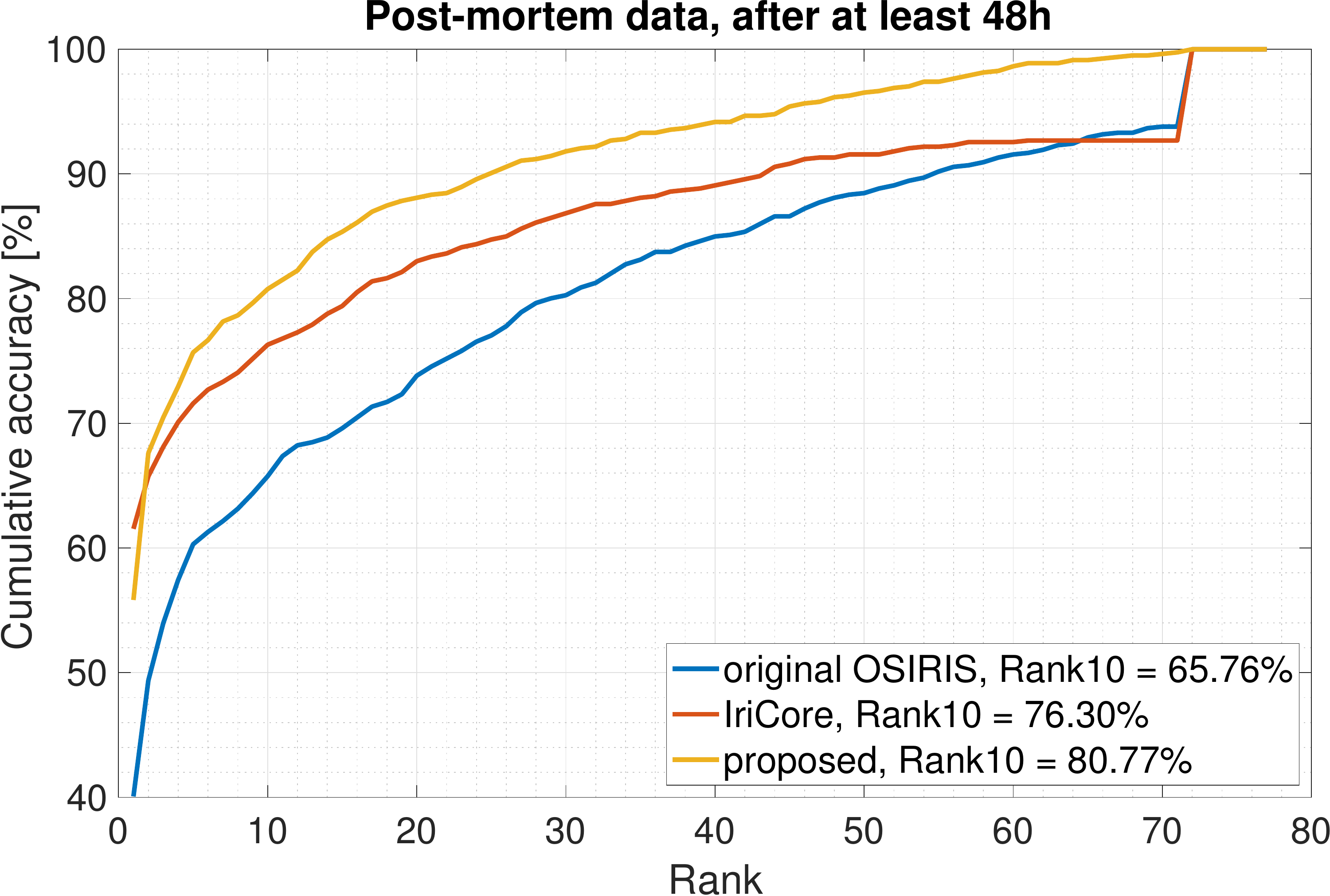}\\\vskip1mm
\includegraphics[width=0.49\linewidth]{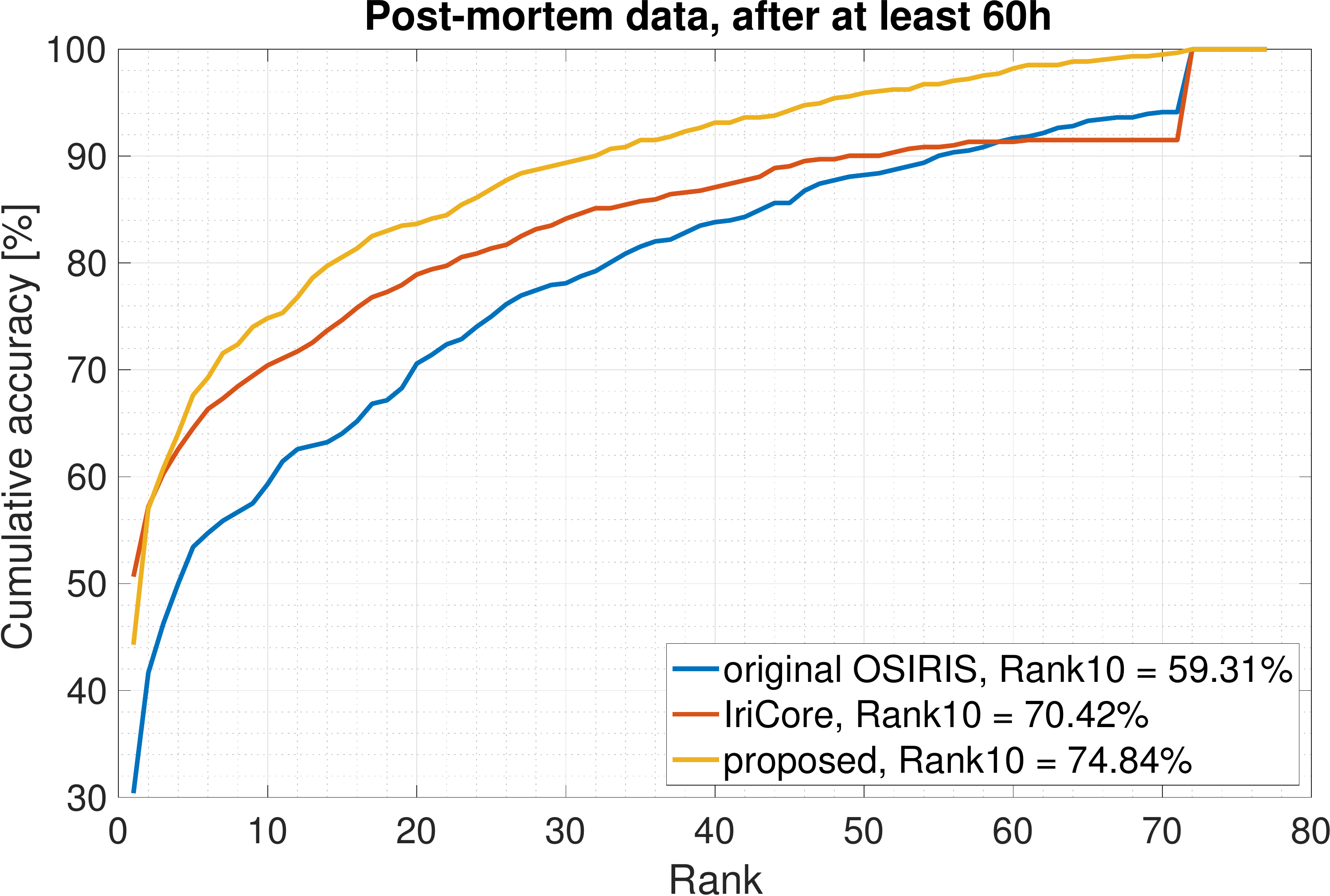}\hskip1mm
\includegraphics[width=0.49\linewidth]{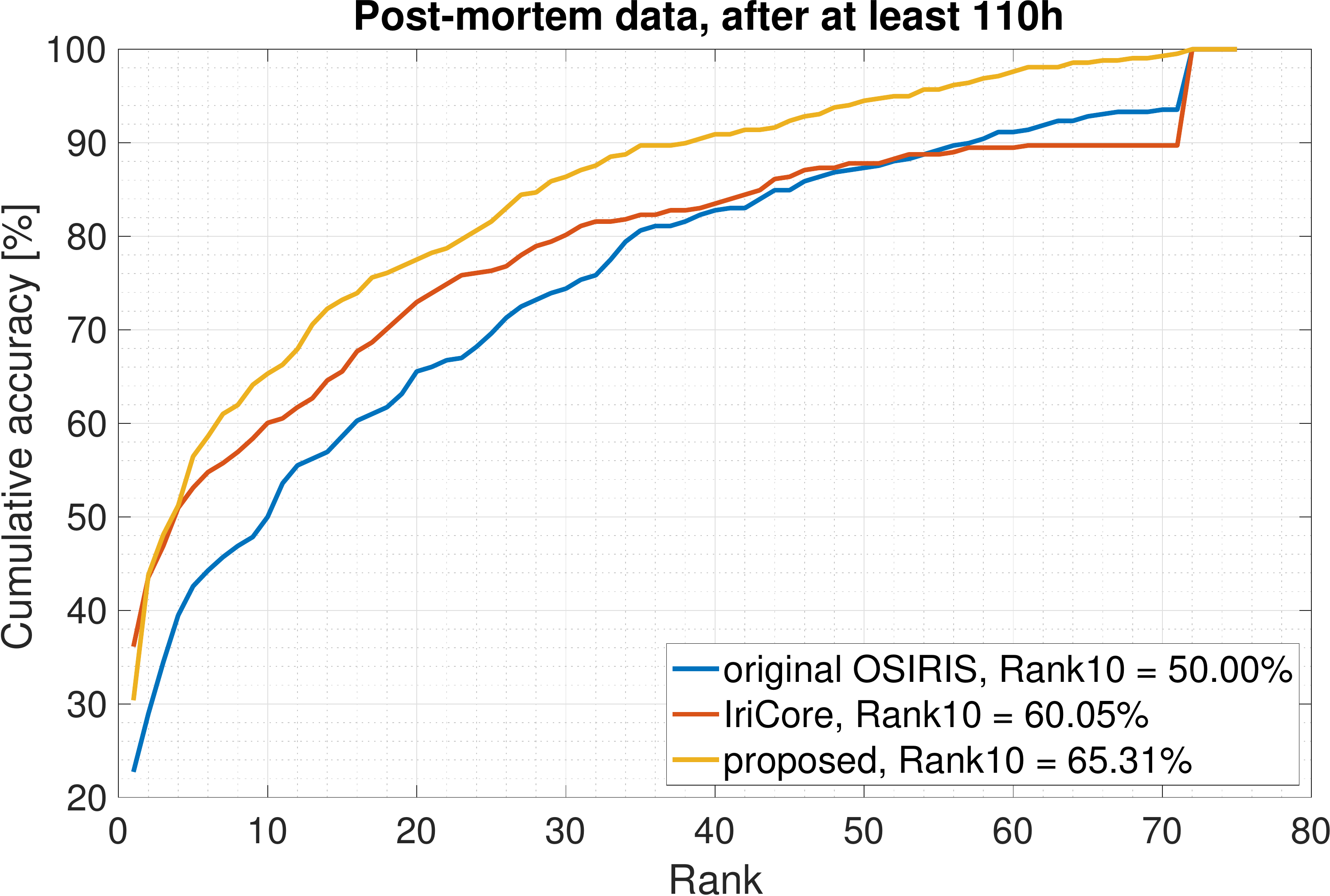}\\\vskip1mm
\includegraphics[width=0.49\linewidth]{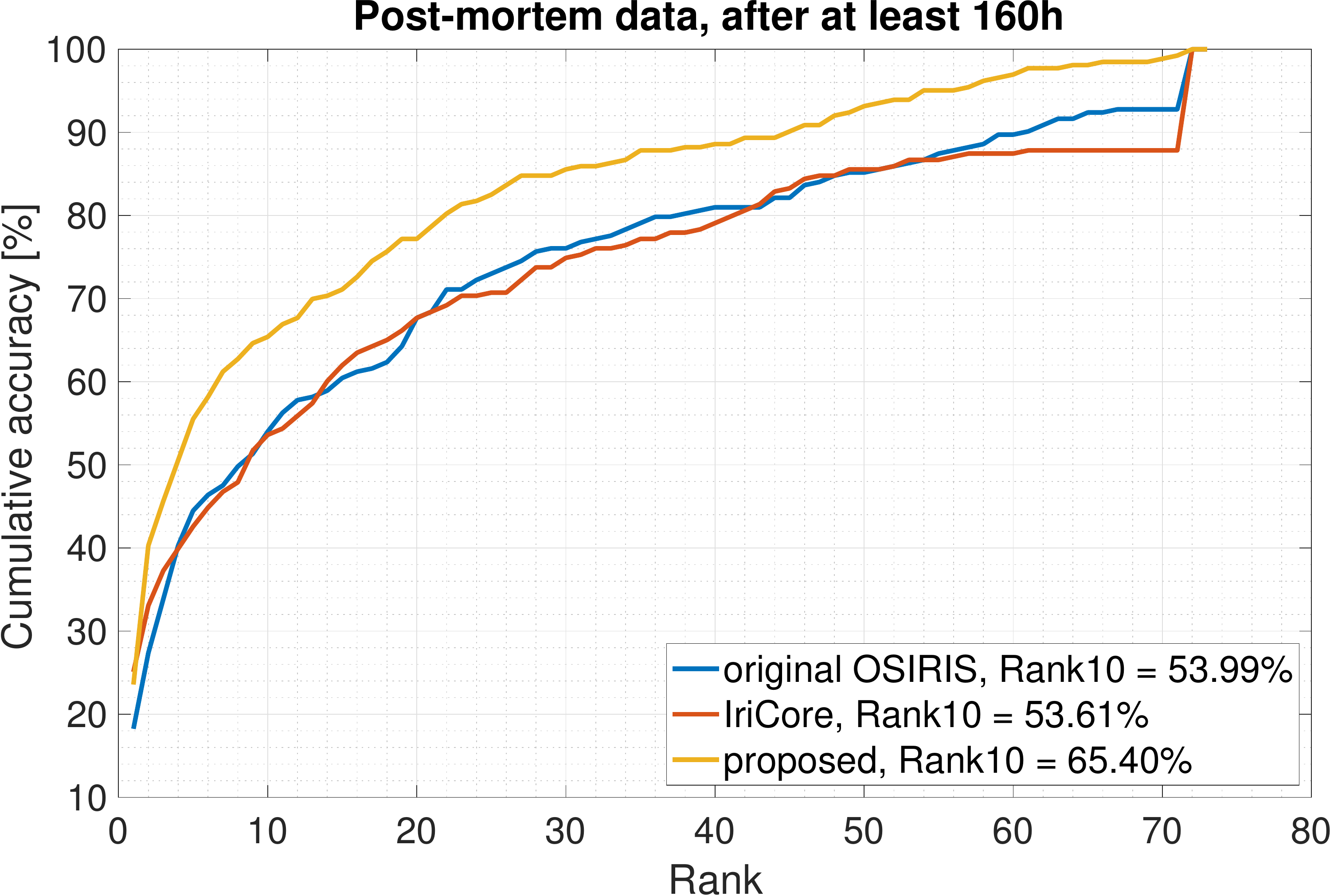}\hskip1mm
\includegraphics[width=0.49\linewidth]{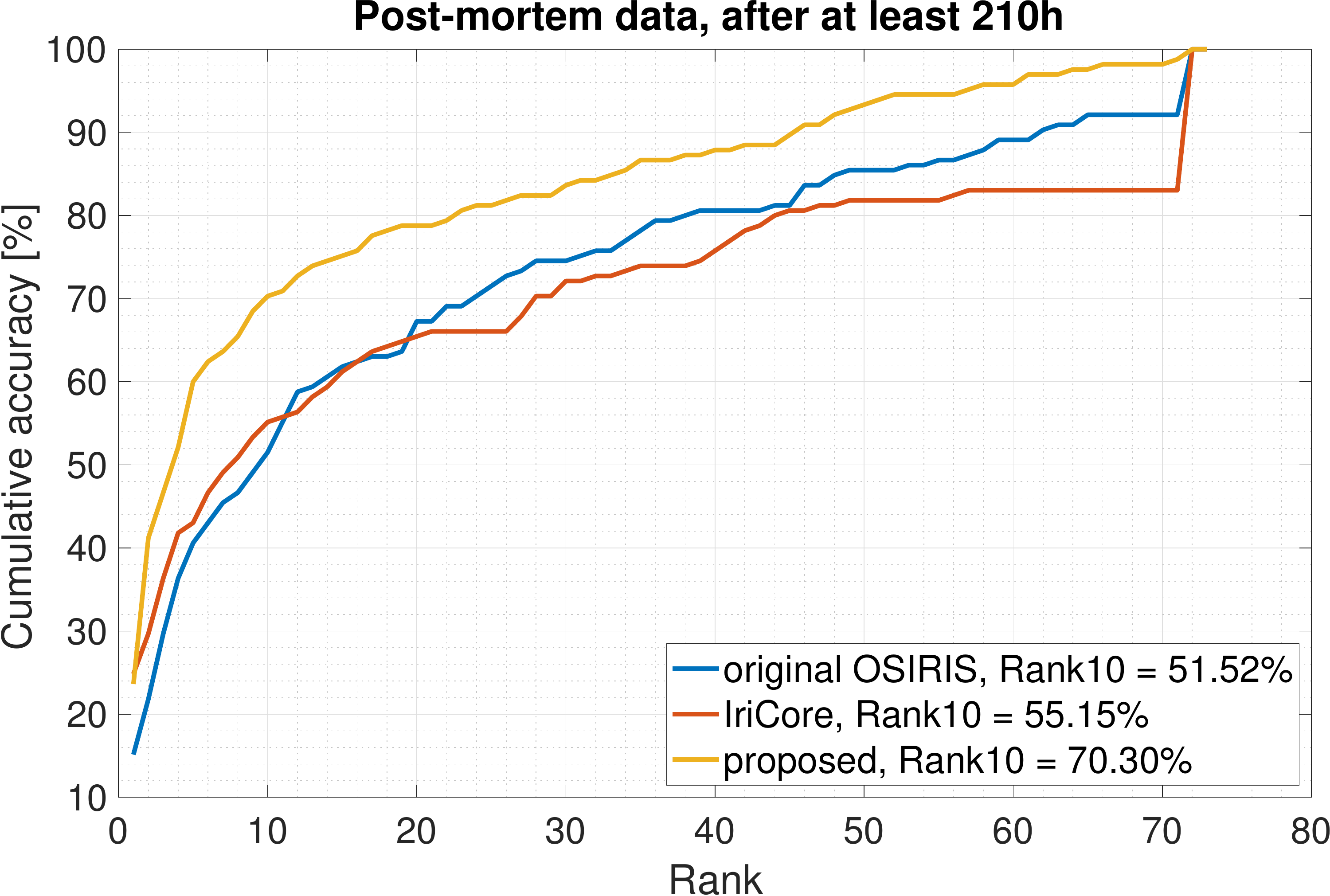}\\
\caption{{Cumulative match characteristic (CMC) curves plotted collectively for six different scenarios of matching the gallery sample (collected shortly after death) with probe samples obtained at least after each of the time periods. Proposed method compared against original OSIRIS method and the commercial matcher. Rank10 scores are given in the plot legends.}}
\label{fig:CMCs}
\end{figure}

{In addition to the verification performance evaluations given by the ROC curves and EER values, we perform the experiment involving candidate list accuracy. Figure \ref{fig:CMCs} presents Cumulative Match Characteristic (CMC) curves for the post-mortem dataset. The experiments for the post-mortem data are constructed in such a way that the gallery sample to which the subsequent probe samples are being matched is always the image collected during the first acquisition session. The probe samples are those collected {\bf at least} after 24 hours post-mortem, 48 hours, 60 hours, 110 hours, 160 hours, and finally after 210 hours, \ie the test subset is getting more difficult in each experiment, as it encompasses progressively more decayed iris samples. 

This experiment can reflect the more real-life application of the method, where the iris recognition algorithm is used as an aid to the forensic examiner, by proposing a candidate list, which is then refined by the human expert. Therefore, the Rank10 values are considered in each experiment. We see that even after 210 hours, or almost 9 days, the chance of including the correct hit on a candidate list comprising 10 suggestions can be as high as 70\%. It has to be noted, though, that the small size of the dataset can cause these error rates to be either under- or over-estimated. The proposed method outperforms both benchmark approaches in the Rank10 metric, with the advantage of our algorithm being the larger the more difficult samples are in the probe set. The Rank10 identification error rates are summarized in Tab. \ref{table:rank10s}}   

\begin{table}[!ht]
\renewcommand{\arraystretch}{1.2}
\caption{{Summary of the Rank10 identification accuracy obtained from the candidate list experiments for the two benchmark methods and the best proposed method ({\bf fine v2highres} model). Numbers of comparisons for each experiment are included in the last column.}}
\label{table:rank10s}
\centering\footnotesize
\begin{tabular}[t]{|c|c|c|c|c|}
\hline
{\bf Probe subset} & {\bf original OSIRIS} & {\bf IriCore} & {\bf proposed} & number of comparisons \\
 & [\%] &[\%]& [\%] & \\
\hline
\hline
{$\geq$ 24h postmortem} & 67.32  & 78.35  & \cellgreen{\bf 83.23} & 240 645 \\
\hline
{$\geq$ 48h postmortem} & 65.76  & 76.30  & \cellgreen{\bf 80.77} & 226 615 \\
\hline
{$\geq$ 60h postmortem} &  59.31 & 70.42 & \cellgreen{\bf 74.84} & 151 280 \\
\hline
{$\geq$ 110h postmortem} & 50.00 & 60.05 & \cellgreen{\bf 65.31} & 138 470 \\
\hline
{$\geq$ 160h postmortem} & 53.99 & 53.61 & \cellgreen{\bf 65.40} & 58 865 \\
\hline
{$\geq$ 210h postmortem} & 51.22 & 55.15 & \cellgreen{\bf 70.30} & 35 685 \\
\hline

\end{tabular}
\end{table}

\subsection{False Non-Match Rate dynamics}
\begin{figure}[h!]
\centering
\includegraphics[width=\linewidth]{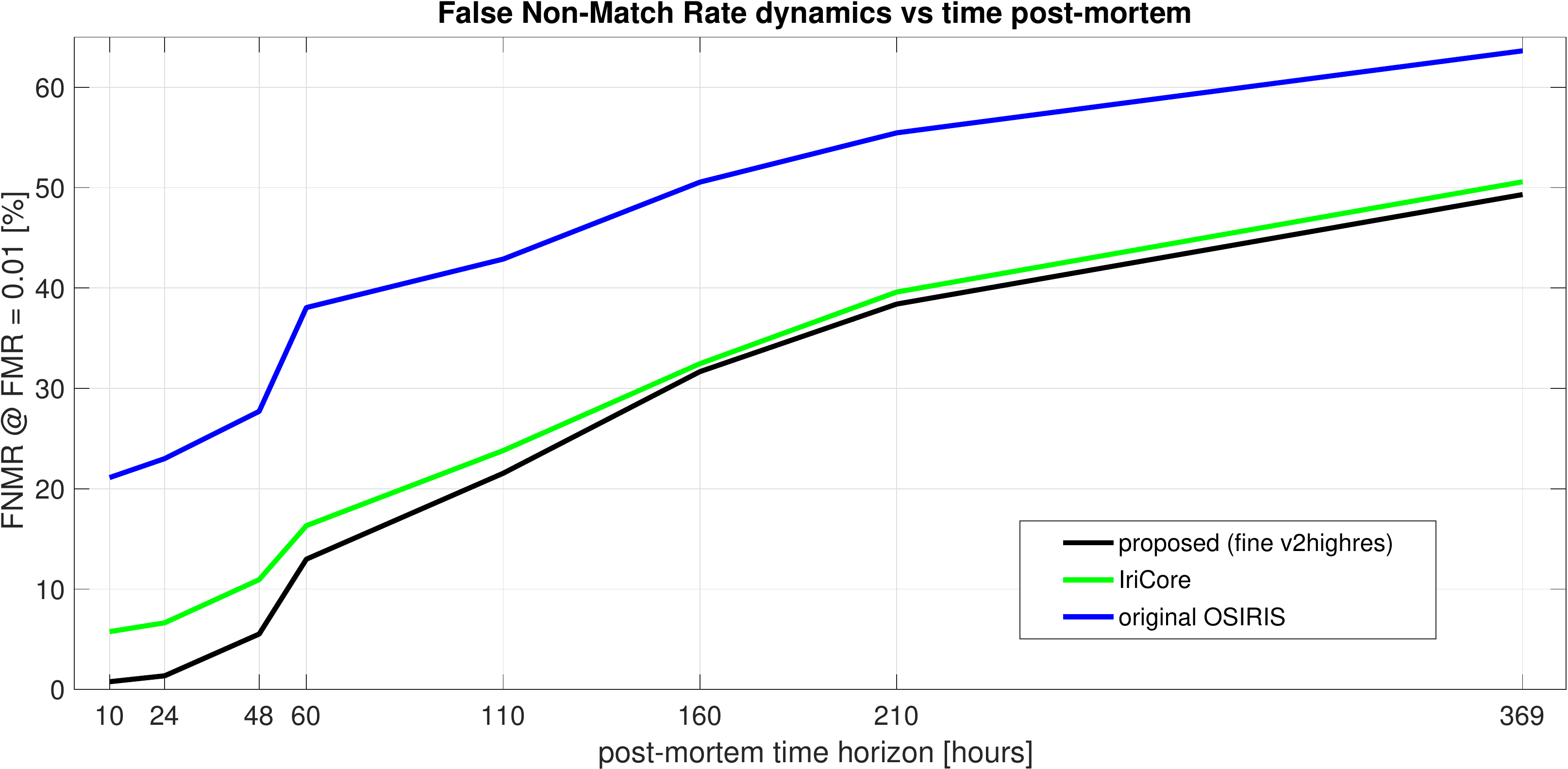}
\caption{Dynamics of False Non-Match Rates in the function of post-mortem sample capture horizon, plotted for two baseline iris recognition methods, and the approach proposed in this paper.}
\label{fig:FNMR}
\end{figure}

In addition to the ROCs, we have also calculated the False Non-Match Rate (FNMR) values at acceptance thresholds which allow the False Match Rate (FMR) values to stay below 1\%. This is done for the post-mortem data to reveal the dynamics of the FNMR as a function of post-mortem sample capture horizon, and therefore to know the chances for a false non-match as time since death progresses. We plot this dynamics for the two baseline methods: OSIRIS and IriCore, as well as the best performing \emph{fine v2highres} DCNN-based modification, Fig. \ref{fig:FNMR}. Notably, for each moment during the increasing post-mortem sample capture time horizon, our proposed approach consistently offers an advantage over the other two algorithms, allowing to reach nearly perfect recognition accuracy for samples collected up to a day after a subject's death.

\section{Conclusions}
\label{sec:Conclusions}
{This work introduces the first known to us attempt at improving iris recognition reliability for post-mortem samples, by offering a robust image segmentation method that can be used as a drop-in image segmentation replacement for the OSIRIS, or any iris recognition pipeline that involves iris normalization onto a dimensionless polar rectangle.

The proposed solution, employing DCNN-based semantic segmentation, allows to achieve close-to-perfect recognition accuracy on the subject-disjoint evaluation database of post-mortem samples, as well as on another challenging database of iris images, collected from elderly individuals with various ophthalmic conditions. This proves the flexibility of the proposed approach in solving problems associated with atypical iris images, which the models introduced in this work can reliably and consistently localize and segment in most of the cases.

Despite the expected drop in recognition accuracy with the increasing time horizon of post-mortem sample collection, the proposed approach consistently outperforms both the academic and the commercial, state-of-the-art iris recognition methods. This in confirmed in both the verification and identification scenario, in which the proposed method was able to offer Rank10 accuracy of more than 70\% when matching probe samples collected 9 days after the gallery sample, compared to only 55\% offered by the open-source matcher and the commercial solution.

The results presented in this work can have great significance for forensics, as they offer substantial improvements in the post-mortem iris matching that can be employed, for instance, for the identification of casualties at crime scenes or mass fatality incidents, as well for ensuring the correctness of cadaver handling. This work directly addresses the current operational trends and needs of forensic practitioners, and we believe it brings valuable contribution to science.}

\section*{Acknowledgment}
The Authors would like to kindly thank Ewelina Bartuzi for her help with generating the CMC curves for the rank list experiments.

\end{document}